\documentclass[runningheads]{llncs}

\usepackage{graphicx}
\usepackage{comment}
\usepackage{lipsum}   %

\usepackage{color,xcolor}
\usepackage{epsfig}

 \usepackage{microtype} 

\usepackage{chngcntr}

\usepackage{tabularx}
\usepackage{adjustbox}
\usepackage{array}
\usepackage{booktabs}
\usepackage{colortbl}
\usepackage{float,wrapfig}
\usepackage{hhline}
\usepackage{multirow}
\usepackage{subcaption}

\usepackage{amsmath,amsfonts,amssymb} %
\usepackage{bm}
\usepackage{nicefrac}
\usepackage{microtype}
\usepackage{dsfont}
\usepackage{changepage}
\usepackage{extramarks}
\usepackage{fancyhdr}
\usepackage{lastpage}
\usepackage{setspace}
\usepackage{soul}
\usepackage{xspace}

\usepackage[pagebackref=true,breaklinks=true,colorlinks,citecolor=black,urlcolor=blue,bookmarks=false]{hyperref}

\usepackage{paralist}
\usepackage{enumitem}

\usepackage{fp}
\usepackage{expl3}[2012-07-08]
\ExplSyntaxOn
\cs_new_eq:NN \fpeval \fp_eval:n
\ExplSyntaxOff

\newcommand{\round}[2]{%
        \FPset{\a}{#1}
        \FPround{\a}{\a}{#2}
        \a
}

\usepackage{amsmath,amsfonts,bm}

\DeclareMathOperator{\trace}{Tr}
\DeclareMathOperator{\diag}{diag}
\newcommand{\dist}{\ell}
\def\x{{x}}
\def\z{{z}}
\def\k{{k}}
\def\v{{v}}
\def\xi{{\x_i}}
\def\zi{{\z_i}}
\def\ki{{\k_i}}
\def\vi{{\v_i}}
\def\xinew{{{\x_*}_i}}
\def\kinew{{{\k_*}_i}}
\def\vinew{{{\v_*}_i}}
\def\thetanew{{\theta_1}}
\def\Lsmooth{{ \mathcal{L}_{\textsf{smooth}}}}
\def\Lconstraint{{ \mathcal{L}_{\textsf{constraint}}}}

\newcommand{\reffig}[1]{Figure~\ref{fig:#1}}
\newcommand{\refsec}[1]{Section~\ref{sec:#1}}
\newcommand{\refapp}[1]{Appendix~\ref{sec:#1}}

\newcommand{\refeq}[1]{Eqn.~\ref{eq:#1}}

\newcommand{\lblsec}[1]{\label{sec:#1}}
\newcommand{\lbleq}[1]{\label{eq:#1}}

\newcommand{\ignorethis}[1]{}

\newcommand{\myparagraph}[1]{\smallskip \noindent \textbf{#1}}

\def\eqref#1{equation~\ref{#1}}

\def\1{\bm{1}}

\newcommand{\test}{\mathcal{D_{\mathrm{test}}}}

\DeclareMathAlphabet{\mathsfit}{\encodingdefault}{\sfdefault}{m}{sl}
\SetMathAlphabet{\mathsfit}{bold}{\encodingdefault}{\sfdefault}{bx}{n}

\newcommand{\E}{\mathbb{E}}

\newcommand{\R}{\mathbb{R}}

\DeclareMathOperator*{\argmin}{arg\,min}

\newcolumntype{L}[1]{>{\raggedright\let\newline\\\arraybackslash\hspace{0pt}}m{#1}}
\newcolumntype{C}[1]{>{\centering\let\newline\\\arraybackslash\hspace{0pt}}m{#1}}
\newcolumntype{R}[1]{>{\raggedleft\let\newline\\\arraybackslash\hspace{0pt}}m{#1}}

\newcommand{\ignore}[1]{}

\makeatletter
\DeclareRobustCommand\onedot{\futurelet\@let@token\@onedot}
\def\@onedot{\ifx\@let@token.\else.\null\fi\xspace}

\makeatother

\definecolor{MyDarkBlue}{rgb}{0,0.08,1}
\definecolor{MyDarkGreen}{rgb}{0.02,0.6,0.02}
\definecolor{MyDarkRed}{rgb}{0.8,0.02,0.02}
\definecolor{MyDarkOrange}{rgb}{0.40,0.2,0.02}
\definecolor{MyPurple}{RGB}{111,0,255}
\definecolor{MyRed}{rgb}{1.0,0.0,0.0}
\definecolor{MyGold}{rgb}{0.75,0.6,0.12}
\definecolor{MyDarkgray}{rgb}{0.66, 0.66, 0.66}

\usepackage[width=122mm,left=12mm,paperwidth=146mm,height=193mm,top=12mm,paperheight=217mm]{geometry}

\begin{document}
\pagestyle{headings}
\mainmatter

\title{Rewriting a Deep Generative Model}

\titlerunning{Rewriting a Deep Generative Model}

\authorrunning{David Bau, Steven Liu, Tongzhou Wang, Jun-Yan Zhu, Antonio Torralba}

\author{David Bau\inst{1}, Steven Liu\inst{1}, Tongzhou Wang\inst{1}, Jun-Yan Zhu\inst{2},  Antonio Torralba\inst{1}}

\institute{MIT CSAIL\inst{1}
\;\;\; Adobe Research\inst{2} \\
}

\maketitle
\begin{abstract}
A deep generative model such as a GAN learns to model a rich set of semantic and physical rules about the target distribution, but up to now, it has been obscure how such rules are encoded in the network, or how a rule could be changed.  In this paper, we introduce a new problem setting:  manipulation of specific rules encoded by a deep generative model. To address the problem, we propose a formulation in which the desired rule is changed by manipulating a layer of a deep network as a linear associative memory. We derive an algorithm for modifying one entry of the associative memory, and we demonstrate that several interesting structural rules can be located and modified within the layers of state-of-the-art generative models.  We present a user interface to enable users to interactively change the rules of a generative model to achieve desired effects, and we show several proof-of-concept applications. Finally, results on multiple datasets demonstrate the advantage of our method against standard fine-tuning methods and edit transfer algorithms. 
\end{abstract}

\section{Introduction}

We present the task of \emph{model rewriting}, which aims to add, remove, and alter the semantic and physical rules of a pretrained deep network.  While modern image editing tools achieve a user-specified goal by manipulating individual input images, we enable a user to synthesize an unbounded number of new images by editing a generative model to carry out modified rules.

\begin{figure}
    \centering
    \includegraphics[width=\textwidth,trim=0 1ex 0 1ex]{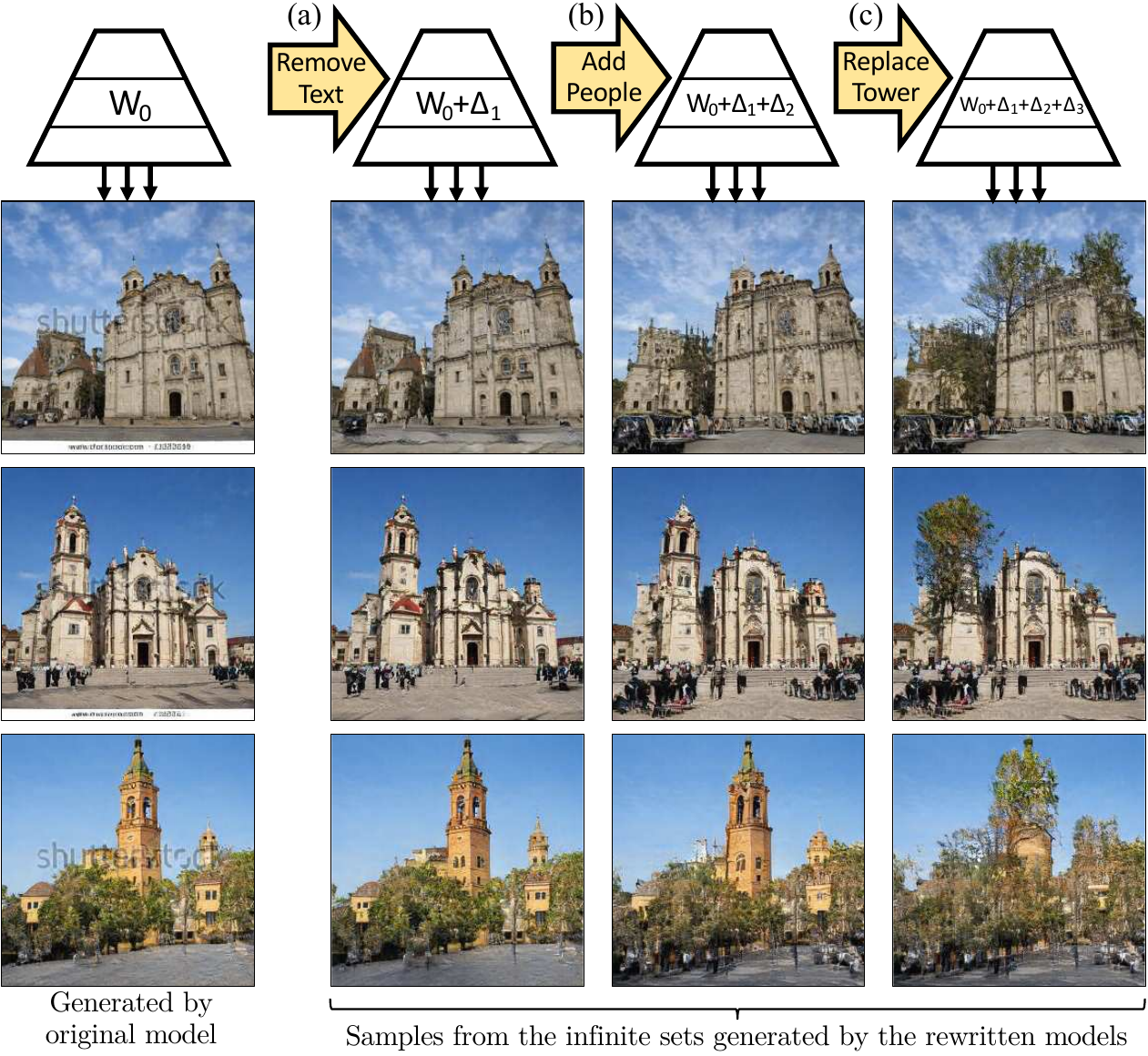}
    \caption{Rewriting the weights of a generator to change generative rules.  Rules can be changed to (a) \emph{remove} patterns such as watermarks; (b) \emph{add} objects such as people; or (c) \emph{replace} definitions such as making trees grow out of towers.  Instead of editing individual images, our method edits the generator, so an infinite set of images can be potentially synthesized and manipulated using the altered rules.}
    \label{fig:teaser}
\end{figure}

For example in \reffig{teaser}, we apply a succession of rule changes to edit a StyleGANv2 model~\cite{karras2020styleganv2} pretrained on LSUN church scenes~\cite{yu2015lsun}.  The first change removes watermark text patterns (a); the second adds crowds of people in front of buildings (b); the third replaces the rule for drawing tower tops with a rule that draws treetops (c), creating a fantastical effect of trees growing from towers.  Because each of these modifications changes the generative model, every single change affects a whole category of images, removing all watermarks synthesized by the model, arranging people in front of many kinds of buildings, and creating tree-towers everywhere. The images shown are samples from an endless distribution.

But why is rewriting a deep generative model useful? A generative model enforces many rules and relationships within the generated images~\cite{bau2019gandissect,jahanian2020steerability}. From a purely scientific perspective, the ability to edit such a model provides insights about what the model has captured and how the model can generalize to unseen scenarios. At a practical level, deep generative models are increasingly useful for image and video synthesis~\cite{mathieu2016deep,zhu2017unpaired,isola2017image,chan2019everybody}. In the future, entire image collections, videos, or virtual worlds could potentially be produced by deep networks, and editing individual images or frames will be needlessly tedious. Instead, we would like to provide authoring tools for modifying the models themselves. With this capacity, a set of similar edits could be {\em transferred} to many images at once.

A key question is how to edit a deep generative model. The computer vision community has become accustomed to training models using large data sets and expensive human annotations, but we wish to enable novice users to easily modify and customize a deep generative model \emph{without} the training time, domain expertise, and computational cost of large-scale machine learning.  In this paper,  we present a new method that can locate and change a specific semantic relationship within a model. In particular, we show how to generalize the idea of a \emph{linear associative memory}~\cite{kohonen1973representation} to a nonlinear convolutional layer of a deep generator. Each layer stores latent rules as a set of key-value relationships over hidden features. Our constrained optimization aims to add or edit one specific rule within the associative memory while preserving the existing semantic relationships in the model as much as possible. We achieve it by directly measuring and manipulating the model's internal structure, without requiring any new training data.

We use our method to create several visual editing effects, including the addition of new arrangements of objects in a scene, systematic removal of undesired output patterns, and global changes in the modeling of physical light. Our method is simple and fast, and it does not require a large set of annotations: a user can alter a learned rule by providing a single example of the new rule or a small handful of examples.  We demonstrate a user interface for novice users to modify specific rules encoded in the layers of a GAN interactively. Finally, our quantitative experiments on several datasets demonstrate that our method outperforms several fine-tuning baselines as well as image-based edit transfer methods, regarding both photorealism and desirable effects. Our code, data, and user interface are available at our \href{rewriting.csail.mit.edu}{website}.  %

\section{Related Work}

\myparagraph{Deep image manipulation.} Image manipulation is a classic problem in computer vision, image processing, and computer graphics. Common operations include color transfer~\cite{reinhard2001color,levin2004colorization}, image deformation~\cite{schaefer2006moving,wolberg1990digital},  object cloning~\cite{perez2003poisson,burt1983laplacian}, and patch-based image synthesis~\cite{efros2001image,barnes2009patchmatch,hertzmann2001image}. %
Recently, thanks to rapid advances of deep generative models~\cite{goodfellow2014generative,kingma2014auto,Hinton06}, learning-based image synthesis and editing methods have become widely-used tools in the community, enabling 
applications such as manipulating the semantics of an input scene~\cite{park2019SPADE,bau2019ganpaint,suzuki2018spatially,collins2020editing},  image colorization~\cite{zhang2016colorful,iizuka2016let,larsson2016learning,zhang2017real}, photo stylization~\cite{gatys2015neural,johnson2016perceptual,luan2017deep,liao2017visual}, image-to-image translation~\cite{isola2017image,zhu2017unpaired,bousmalis2017unsupervised,taigman2017unsupervised,liu2017unsupervised,huang2018multimodal}, and face editing and synthesis~\cite{fried2019text,nagano2018pagan,Faceshop}. While our user interface is inspired by previous interactive systems, our goal is {\em not} to manipulate and synthesize a single image using deep models. Instead, our work aims to manipulate the structural rules of the model itself, creating an altered deep network that can produce countless new images following the modified rules. 

\myparagraph{Edit transfer and propagation.} Edit transfer methods propagate pixel edits to corresponding locations in other images of the same object or adjacent frames in the same video~\cite{an2008appprop,xu2009efficient,hasinoff2010search,chen2012manifold,chen2014sparse,yucer2012transfusive,endo2016deepprop}. These methods achieve impressive results but are limited in two ways. First, they can only transfer edits to images of the same instance, %
as image alignment between different instances is challenging. Second, the edits are often restricted to color transfer or object cloning. In contrast, %
our method can change context-sensitive rules that go beyond pixel correspondences (\refsec{exp-reflection}).  In~\refsec{exp-paste}, we compare to an edit propagation method based on state-of-the-art alignment algorithm, Neural Best-Buddies~\cite{aberman2018neural}.

\myparagraph{Interactive machine learning} systems aim to improve training through human interaction in labeling~\cite{cohn1994improving,fails2003interactive,settles2008analysis}, or by allowing a user to to aid in the model optimization process via interactive feature selection~\cite{dy2000visualization,guo2003coordinating,raghavan2006active,krause2014infuse} or model and hyperparameter selection~\cite{kapoor2010interactive,patel2011using,jiang2017interactive}. Our work differs from these previous approaches because rather than asking for human help to attain a fixed objective, we enable a user to solve novel creative modeling tasks, given a pre-trained model. Model rewriting allows a user to create a network with new rules that go beyond the patterns present in the training data.

\myparagraph{Transfer learning and model fine-tuning.} Transfer learning adapts a learned model to unseen learning tasks, domains, and settings. Examples include domain adaptation~\cite{saenko2010adapting}, zero-shot or few-shot learning~\cite{socher2013zero,lake2015human}, model pre-training and feature learning~\cite{donahue2014decaf,zeiler2014visualizing,yosinski2014transferable}, and meta-learning~\cite{bengio1992optimization,andrychowicz2016learning,finn2017model}. Our work differs because instead of extending the training process with more data or annotations, we enable the user to directly change the behavior of the existing model through a visual interface. Recently, several methods~\cite{ulyanov2018deep,shocher2018zero,bau2019ganpaint} propose to train or fine-tune an image generation model to a particular image for editing and enhancement applications. Our goal is different, as we aim to identify and change rules that can generalize to many different images instead of one.

\section{Method}
\lblsec{method}
To rewrite the rules of a trained generative model, we allow users to specify a handful of model outputs that they wish to behave differently. Based on this objective, we optimize an update in model weights that generalizes the requested change. In \refsec{method}, we derive and discuss this optimization. In \refsec{ui}, we present the user interface that allows the user to interactively define the objective and edit the model.

\refsec{objective} formulates our objective on how to add or modify a specific rule while preserving existing rules. We then consider this objective for linear systems and connect it to a classic technique---associative memory~\cite{kohonen1972correlation,anderson1972simple,kohonen2012associative} (\refsec{associate}); this perspective allows us to derive a simple update rule (\refsec{update}). Finally, we apply the solution to the nonlinear case and derive our full algorithm (\refsec{nonlinear}).

\subsection{Objective: Changing a Rule with Minimal Collateral Damage}
\lblsec{objective}
Given a pre-trained generator $G(\z; \theta_0)$ with weights $\theta_0$, we can synthesize multiple images $\xi = G(\zi; \theta_0)$, where each image is produced by a latent code $\zi$. 
Suppose we have manually created desired changes $\xinew$ for those cases. 
We would like to find updated weights $\theta_1$ that change a computational rule to match our target examples $\xinew  \approx G(\zi; \thetanew)$, while minimizing interference with other behavior:
\begin{align}
\theta_1 & = \arg\min_{\theta} \Lsmooth(\theta) + \lambda \Lconstraint(\theta), \\ \Lsmooth(\theta) & \triangleq   \E_{\z}\left[ \dist( G(\z; \theta_0) , G(\z; \theta) ) \right],  \\
\Lconstraint(\theta) & \triangleq \sum_i \dist( \xinew,  G(\zi; \theta) ).
\lbleq{ideal-constraint} 
\end{align}
A traditional solution to the above problem is to jointly optimize the weighted sum of $\Lsmooth$ and $\Lconstraint$ over $\theta$, where  $\dist( \cdot )$ is a distance metric that measures the perceptual distance between images~\cite{johnson2016perceptual,dosovitskiy2016generating,zhang2018unreasonable}. 
Unfortunately, this standard approach does not produce a generalized rule within $G$, because the large number of parameters $\theta$ allow the generator to quickly overfit the appearance of the new examples without good generalization; we evaluate this approach in \refsec{expr}.

However, the idea becomes effective with two modifications: (1) instead of modifying all of $\theta$, we reduce the degrees of freedom by modifying weights $W$ at only one layer, and (2) for the objective function,
we directly minimize distance in the output feature space of that same layer.

Given a layer $L$, we use $\k$ to denote the features computed by the first $L-1$ fixed layers of $G$, and then write $\v=f(\k;W_0)$ to denote the computation of layer $L$ itself, with pretrained weights $W_0$.  For each exemplar latent $\zi$, these layers produce features $\kinew$ and $\vi = f(\kinew; W_0)$.  Now suppose, for each target example $\xinew$, the user has manually created a feature change $\vinew$.  (A user interface to create target feature goals is discussed in \refsec{ui}.)  Our objective becomes:
\begin{align}
W_1 & = \arg\min_{W} \Lsmooth(W) + \lambda \Lconstraint(W), \\ \Lsmooth(W) & \triangleq   \E_{\k}\left[\; || f(\k; W_0) - f(\k; W) ||^2 \; \right],  \\
\Lconstraint(W) & \triangleq \sum_i || \vinew -  f(\kinew; W)||^2,
\lbleq{lconstraint}
\end{align}
where $||\cdot||^2$ denotes the L2 loss. Even within one layer, the weights $W$ contain many parameters. But the degrees of freedom can be further reduced to constrain the change to a specific direction that we will derive;  this additional directional constraint will allow us to create a generalized change from a single $(k_*, v_*)$ example.  To understand the constraint, it is helpful to interpret a single convolutional layer as an associative memory, a classic idea that we briefly review next.

\subsection{Viewing a Convolutional Layer as an Associative Memory}
\lblsec{associate}
Any matrix $W$ can be used as an associative memory~\cite{kohonen2012associative} that stores a set of key-value pairs $\{(\ki, \vi)\}$ that can be retrieved by matrix multiplication:
\begin{align}
\vi \approx W \ki. 
\lbleq{linearmap}
\end{align}
The use of a matrix as a \emph{linear associative memory} is a foundational idea in neural networks~\cite{kohonen1972correlation,anderson1972simple,kohonen2012associative}. For example, if the keys $\{\ki\}$ form a set of mutually orthogonal unit-norm vectors, then an error-free memory can be created as
\begin{align}
W_{\text{orth}} \triangleq \sum_{i} \vi \ki^T.
\end{align}
Since $\ki^T \k_j = 0$ whenever $i \neq j$, all the irrelevant terms cancel when multiplying by $\k_j$, and we have $W_{\text{orth}} \; \k_j = \v_j$. A new value can be stored by adding $\v_* {\k_*}^T$ to the matrix as long as $\k_*$ is chosen to be orthogonal to all the previous keys. This process can be used to store up to $N$ associations in an $M\times N$ matrix.
\begin{figure}[t]
\centering
\includegraphics[width=1.0\textwidth,trim=0 1ex 0 1ex]{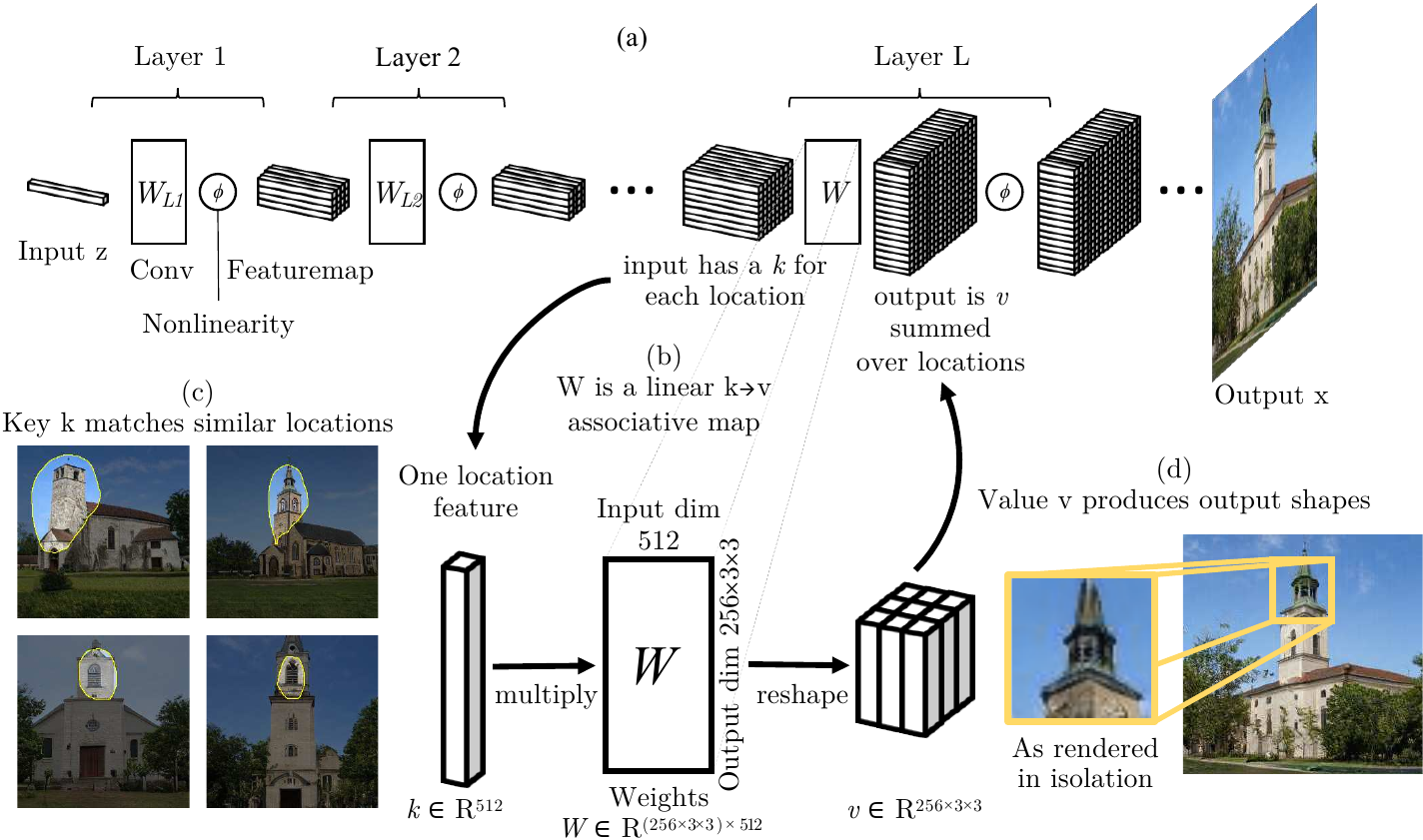}
\caption{(a) A generator consists of a sequence of layers; we focus on one particular layer $L$. (b) The convolutional weights $W$ serve an associative memory, mapping keys $k$ to values $v$.  The keys are single-location input features, and the values are patterns of output features.  (c) A key will tend to match semantically similar contexts in different images. Shown are locations of generated images that have features that match a specific $k$ closely. (d) A value renders shapes in a small region.  Here the effect of a value $v$ is visualized by rendering features at one location alone, with features at other locations set to zero.  Image examples are taken from a StyleGANv2 model trained on LSUN outdoor church scenes.}
    \label{fig:keys-values}
\end{figure}

Figure~\ref{fig:keys-values} views the weights of one convolutional layer in a generator as an associative memory.  Instead of thinking of the layer as a collection of convolutional filtering operations, we can think of the layer as a memory that associates keys to values.  
Here each key $\k$ is a single-location feature vector. The key is useful because, in our trained generator, the same key will match many semantically similar locations across different images, as shown in \reffig{keys-values}c.  Associated with each key, the map stores an output value $\v$ that will render an arrangement of output shapes. This output can be visualized directly by rendering the features in isolation from neighboring locations, as shown in \reffig{keys-values}d. 

For example, consider a layer that transforms a 512-channel featuremap into a 256-channel featuremap using a $3\times 3$ convolutional kernel; the weights form a $256\times512\times3\times3$ tensor.
For each key $k\in \R^{512}$, our layer will recall a value $v \in \R^{256\times 3\times 3} = \R^{2304}$ representing a $3\times3$ output pattern of $256$-channel features, flattened to a vector, as $v=Wk$.  Our interpretation of the layer as an associative memory does not change the computation: the tensor is simply reshaped and treated as a dense rectangular matrix $W \in \R^{(256 \times 3 \times 3) \times 512}$, whose job is to map keys $k\in \R^{512}$ to values $v\in\R^{2304}$, via \refeq{linearmap}.

\textbf{Arbitrary Nonorthogonal Keys.}
In classic work, Kohonen~\cite{kohonen1973representation} observed that an associative memory can support more than $N$ nonorthogonal keys $\{\ki\}$ if instead of requiring exact equality $\vi = W \ki$, we choose $W_0$ to minimize error:
\begin{align}
W_0 \triangleq \argmin_{W} \sum_i  \, || \vi - W \ki ||^2.
\lbleq{ls-objective}
\end{align}
To simplify notation, let us assume a finite set of pairs $\{(\ki, \vi)\}$ and collect keys and values into matrices $K$ and $V$ whose $i$-th column is the $i$-th key or value:
\begin{align}
K &\triangleq \left[ k_1 | k_2 | \cdots | k_i | \cdots \right], \\
V &\triangleq \left[ v_1 | v_2 | \cdots | v_i | \cdots \right].
\end{align}
The minimization (\refeq{ls-objective}) is the standard linear least-squares problem.  A unique minimal solution can be found by solving for $W_0$ using the normal equation $W_0 K K^T = V K^T$, or equivalently by using the pseudoinverse $W_0 = V K^+$. 

\subsection{Updating $W$ to Insert a New Value}
\lblsec{update}
Now, departing from Kohonen~\cite{kohonen1973representation}, we ask how to modify $W_0$.  Suppose we wish to overwrite a single key to assign a new value $k_* \rightarrow v_*$ provided by the user.  After this modification, our new matrix $W_1$ should satisfy two conditions:
\begin{align}
W_1 & = \argmin_W || V - W K ||^2,  \lbleq{cls-min} \\
\text{subject to} \; v_* & = W_1 k_*. \lbleq{cls-const} 
\end{align}
That is, it should store the new value; and it should continue to minimize error in all the previously stored values. This forms a constrained linear least-squares (CLS) problem which can be solved exactly as $W_1 K K^T = V K^T + \Lambda\,{k_*}^T$, where the vector $\Lambda \in \R^m$ is determined by solving the linear system with the constraint in \refeq{cls-const} (see \refapp{solving-lambda}).  Because $W_0$ satisfies the normal equations, we can expand $V K^T$ in the CLS solution and simplify:
\begin{align}
W_1 K K^T & = W_0 K K^T + \Lambda\,{k_*}^T \lbleq{wkk}
\\
W_1 & = W_0 + \Lambda (C^{-1}k_*)^{T}  \lbleq{rank1-update}
\end{align}
Above, we have written $C \triangleq K K^T$ as the second moment statistics.  ($C$ is symmetric; if $K$ has zero mean, $C$ is the covariance.) 
Now \refeq{rank1-update} has a simple form.  Since $\Lambda\in\R^m$ and $(C^{-1}k_*)^{T} \in \R^n$ are simple vectors, the update $\Lambda (C^{-1}k_*)^{T}$ is a rank-one matrix with rows all multiples of the vector $(C^{-1}k_*)^T$.

\refeq{rank1-update} is interesting for two reasons.  First, it shows that enforcing the user's requested mapping $k_* \rightarrow v_*$ transforms the soft error minimization objective (\ref{eq:cls-min}) into the hard constraint that the weights be updated in a particular straight-line direction $C^{-1}k_*$.  Second, it reveals that the update direction is determined only by the overall key statistics and the specific targeted key $k_*$. The covariance $C$ is a model constant that can be pre-computed and cached, and the update direction is determined by the key \emph{regardless of any stored value}. Only $\Lambda$, which specifies the magnitude of each row change, depends on the target value $v_*$.

\subsection{Generalize to a Nonlinear Neural Layer}
\lblsec{nonlinear}
In practice, even a single network block contains several non-linear components such as a biases, ReLU, normalization, and style modulation. Below, we generalize our procedure to the nonlinear case where the solution to $W_1$ cannot be calculated in a closed form. We first define our update direction: 
\begin{align}
d \triangleq C^{-1}k_*.
\lbleq{d-def}
\end{align}
Then suppose we have a non-linear neural layer $f(k; W)$ which follows the linear operation $W$ with additional nonlinear steps.   Since the form of \refeq{rank1-update} is sensitive to the rowspace of $W$ and insensitive to the column space, we can use the same rank-one update form to constrain the optimization of $f(k_*; W) \approx v_*$.

Therefore, in our experiments, when we update a layer to insert a new key $k_* \rightarrow v_*$, we begin with the existing $W_0$, and we perform an optimization over the rank-one subspace defined by the row vector $d^T$ from \refeq{d-def}.  That is, in the nonlinear case, we update $W_1$  by solving the following optimization:
\begin{align}
\Lambda_1 = \argmin_{\Lambda \in \R^M} || v_* - f(k_*; W_0 + \Lambda \,d^T) ||.
\lbleq{nl-ins-rule}
\end{align}
Once $\Lambda_1$ is computed, we update the weight as $W_1 =  W_0 + \Lambda_1 d^T$. %

Our desired insertion may correspond to a change of more than one key at once, particularly if our desired target output forms a feature map patch $V_*$ larger than a single convolutional kernel, i.e., if we wish to have $V_* = f(K_*; W_1)$ where $K_*$ and $V_*$ cover many pixels. To alter $S$ keys at once, we can define the allowable deltas as lying within the low-rank space spanned by the $N\times S$ matrix $D_S$ containing multiple update directions $d_i = C^{-1}K_{*i}$, indicating which entries of the associative map we wish to change.
\begin{align}
\Lambda_S &= \argmin_{\Lambda \in \R^{M\times S}} || V_* - f(K_*; W_0 + \Lambda\,{D_S}^T) ||, \lbleq{multidim-constrained-loss}
\\
\text{where}\;D_S &\triangleq \left[ d_{1} | d_{2} | \cdots | d_{i} | \cdots | d_{S} \right].
\lbleq{dset-def}
\end{align}
We can then update the layer weights using $W_S =  W_0 + \Lambda_S {D_S}^T$. The change can be made more specific by reducing the rank of $D_S$; details are discussed \refapp{rank-reduction}.
To directly connect this solution to our original objective (\refeq{lconstraint}), we note that the constrained optimization can be solved using projected gradient descent.  That is, we relax \refeq{multidim-constrained-loss} and use optimization to minimize $\argmin_W ||V_* - f(K_*; W)||$; then, to impose the constraint, after each optimization step, project $W$ into into the subspace $W_0 + \Lambda_S {D_S}^T$.

\begin{figure}[t]
\centering
\includegraphics[width=0.9\textwidth,trim=0 1ex 0 1ex]{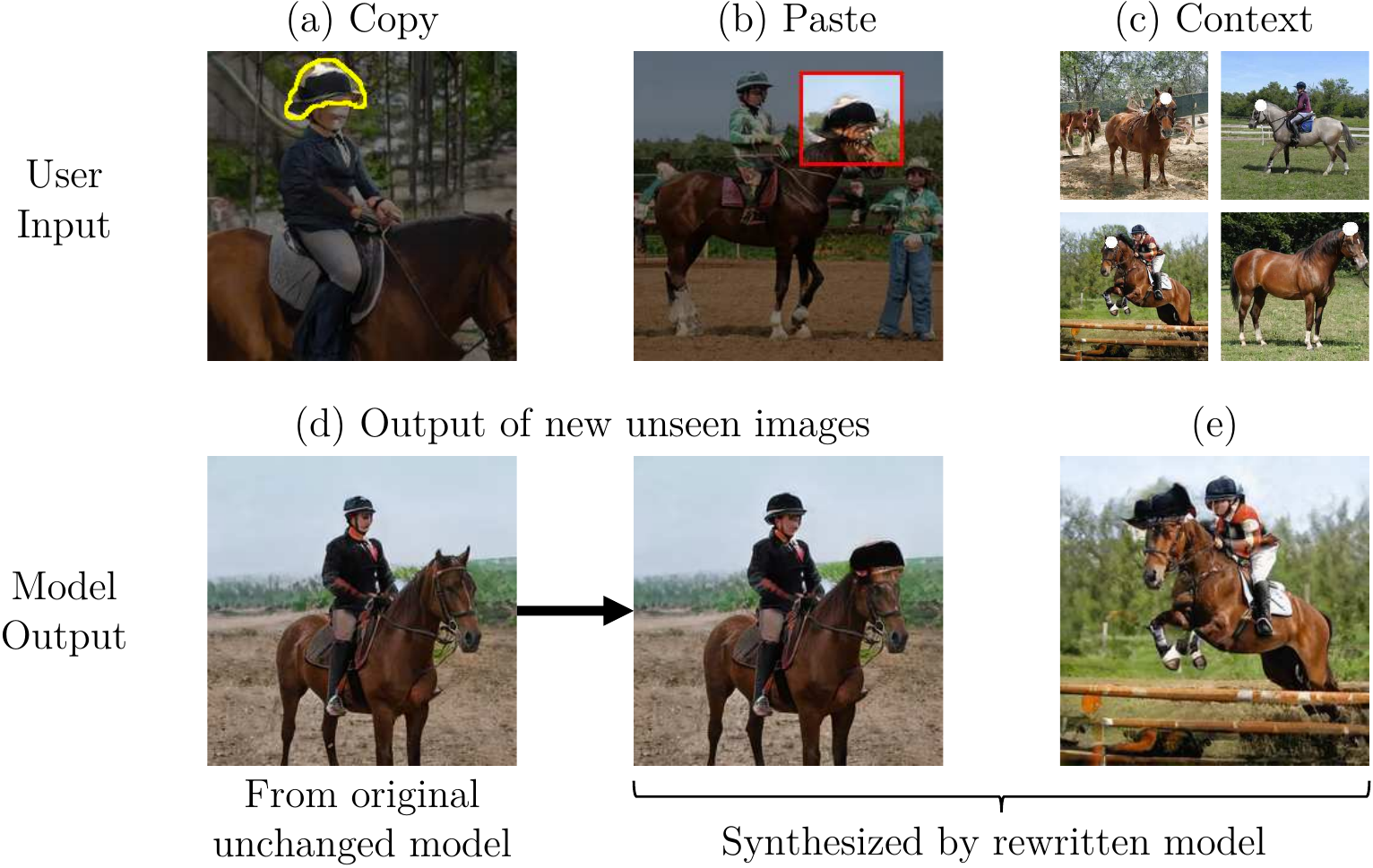}
\caption{The \emph{Copy-Paste-Context} interface for rewriting a model.  (a) \textbf{Copy:} the user uses a brush to select a region containing an interesting object or shape,
defining the target value $V_*$.  
(b) \textbf{Paste:}  The user positions and pastes the copied object into a single target image. This specifies the $K_* \rightarrow V_*$ pair constraint.
(c) \textbf{Context:} To control generalization, the user selects target regions in several images. This establishes the updated direction $d$ for the associative memory. 
(d) The edit is applied to the model, not a specific image, so newly generated images will always have hats on top of horse heads. 
(e) The change has generalized to a variety of different types of horses and poses (see more in \refapp{more-examples}).}
\label{fig:edit-interface}
\end{figure}
\section{User Interface}
\lblsec{ui}
To make model rewriting intuitive for a novice user, we build a user interface that provides a three-step rewriting process: \emph{Copy}, \emph{Paste}, and \emph{Context}. %

\myparagraph{Copy and Paste} allow the user to copy an object from one generated image to another. The user browses through a collection of generated images and highlights an area of interest to copy; then selects a generated target image and location for pasting the object. For example,  in \reffig{edit-interface}a, the user selects a helmet worn by a rider and then pastes it in \reffig{edit-interface}b on a horse's head.

Our method downsamples the user's copied region to the resolution of layer $L$ and gathers the copied features as the target value $V_*$.  Because we wish to change not just one image, but the model rules themselves, we treat the pasted image as a new rule $K_* \rightarrow V_*$ associating the layer $L-1$ features $K_*$ of the target image with the newly copied layer $L$ values $V_*$ that will govern the new appearance.

\myparagraph{Context Selection} allows a user to specify how this change will be generalized, by pointing out a handful of similar regions that should be changed.  
For example, in \reffig{edit-interface}b, the user has selected heads of different horses. %

We collect the layer $L-1$ features at the location of the context selections as a set of relevant $K$ that are used to determine the weight update direction $d$ via \refeq{d-def}. Generalization improves when we allow the user to select several context regions to specify the update direction (see Table~\ref{tab:ffhq}); in \reffig{edit-interface}, the four examples are used to create a single $d$.  \refapp{rank-reduction} discusses this rank reduction.

Applying one rule change on a StyleGANv2 model requires about eight seconds on a single Titan GTX GPU. Please check out the demo \href{http://rewriting.csail.mit.edu/video}{video} of our interface. %

\section{Results}
\lblsec{expr}
We test model rewriting with three editing effects. First, we add new objects into the model, comparing results to several baseline methods. Then, we use our technique to erase objects using a low-rank change; we test this method on the challenging watermark removal task.  Finally, we invert a rule for a physical relationship between bright windows and reflections in a model.

\begin{figure}
    \centering
    \includegraphics[width=0.92\textwidth]{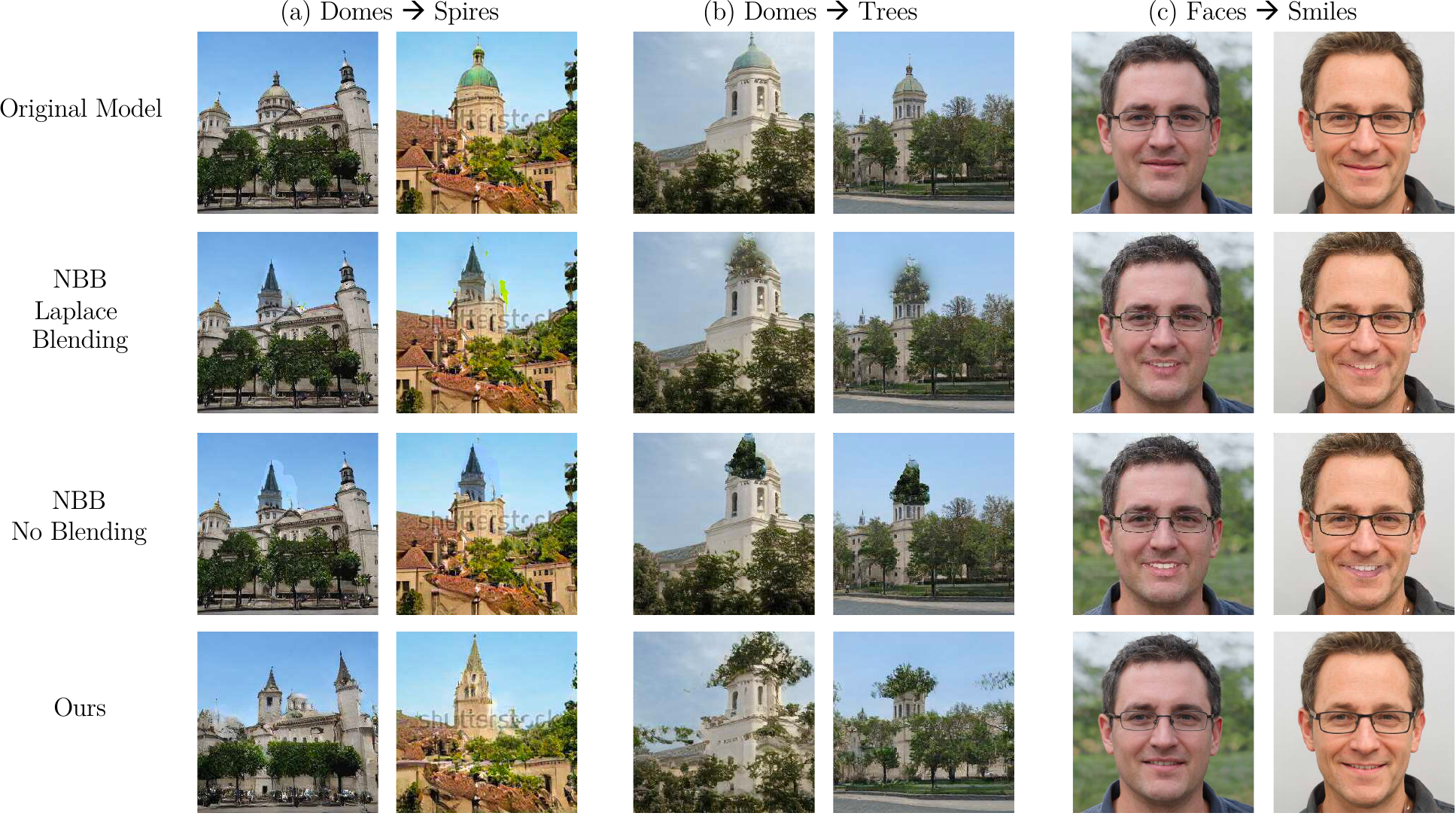}
    \caption{Adding and replacing objects in three different settings.  (a) Replacing domes with an angular peaked spire causes peaked spires to be used throughout the model.  (b) Replacing domes with trees can generate images unlike any seen in a training set. (c) Replacing closed lips with an open-mouth smile produces realistic open-mouth smiles.  For each case, we show the images generated by an unchanged model, then the edit propagation results, with and without blending. Our method is shown in the last row.}
    \label{fig:paste-additions}
\end{figure}
\subsection{Putting objects into a new context}
\lblsec{exp-paste}
Here we test our method on several specific model modifications.  In a church generator, the model edits change the shape of domes to spires, and change the domes to trees, and in a face generator, we add open-mouth smiles.  Examples of all the edits are shown in \reffig{paste-additions}.

\begin{table}
    \newcommand{\mround}[1]{\round{#1}{2}}
    \newcommand{\na}{\multicolumn{1}{c}{--}}
    \centering \small
        \resizebox{0.8\width}{!}{%
    \begin{tabular}{l c c c}
    & \% smiling images $\uparrow$
    & LPIPS (masked) $\downarrow$
    & \shortstack{\% more realistic \\ than ours $\uparrow$} 
    \\ \hline

    Our method (projected gradient descent)
    & 84.37
    & \textbf{\mround{0.035084010987915096}}
    & \na
    \\
    
    With direct optimization of $\Lambda$
    & 87.44
    & \mround{0.1375907236509025}
    & 43.0 %
    \\

    With single-image direction constraint
    & 82.12
    & \mround{0.050821064344234765}
    & 47.3 %
    \\
    
    With single-layer, no direction constraint 
    & 90.94
    & \mround{0.304}
    & 6.8 %
    \\
    \hline

    Finetuning all weights
    & 85.78
    & \mround{0.40296847543120384}
    & 8.7 %
    \\
    \hline
    
    NBB + Direct copying
    & 94.81
    & \mround{0.3241583276927471}
    & 9.8 %
    \\

    NBB + Laplace blending
    & \textbf{93.51}
    & \mround{0.32196807104349134}
    & 8.6 %
    \\
    \hline
    
    Unmodified model 
    & 78.37
    & \na
    & 50.9 %
    \\
    \end{tabular}
    }

    \caption{Editing a StyleGANv2 \cite{karras2020styleganv2} FFHQ \cite{karras2019style} model to produce smiling faces in  $n=10,000$ images. To quantify the efficacy of the change, we show the percentage of smiling faces among the modified images, and we report the LPIPS distance on masked images to quantify undesired changes.  For realism, workers make $n=1,000$ pairwise judgements comparing images from other methods to ours.}
    \label{tab:ffhq}
\end{table}

\begin{table}
    \newcommand{\mround}[1]{\round{#1}{2}}
    \newcommand{\nastar}{\multicolumn{1}{c}{\hspace{15pt}--~*}}
    \newcommand{\na}{\multicolumn{1}{c}{--}}
    \centering \small
    
    {
        \centering \small
        
        \resizebox{
          \ifdim\width>\textwidth
            \textwidth
          \else
            \width
          \fi
        }{!}{%
        \begin{tabular}{l @{\extracolsep{5pt}} c c c c c @{}}
        & \multicolumn{3}{c}{Dome $\rightarrow$ Spire}
        & \multicolumn{2}{c}{Dome $\rightarrow$ Tree}
        \\
        \cline{2-4} \cline{5-6}
        & \shortstack{\% \texttt{dome} pixels\\correctly modified} $\uparrow$
        & \shortstack{LPIPS\\(masked) $\downarrow$}
        & \shortstack{\% more realistic \\ than ours $\uparrow$} 
        & \shortstack{\% \texttt{dome} pixels\\correctly modified} $\uparrow$
        & \shortstack{LPIPS\\(masked) $\downarrow$}
        \\ \hline
        
        Our method (projected gradient descent)
        & \textbf{\mround{92.03}} &
        \textbf{\mround{0.020296541112358683}}
        & \na
        &{\mround{48.65}} &
        \textbf{\mround{0.02591632406162098}}
        \\
        
        With direct optimization of $\Lambda$ 
        & \mround{80.03134339237554}
        & \mround{0.10131881302036345}
        & 53.7
        & \textbf{\mround{59.42852977622276}}
        & \mround{0.13102583499923348}
        \\

        With single-image direction constraint 
        & \mround{90.14} 
        & \mround{0.039834109118301424}
        & 48.8 %
        & \mround{39.72}
        & \mround{0.02659131222628057}
        \\
        
        With single-layer, no direction constraint
        & \mround{80.69}
        & \mround{0.294}
        & 38.1
        & \mround{41.32}
        & \mround{0.445}
        \\
        \hline
        
        Finetuning all weights
        & \mround{41.157672105}
        & \mround{0.3622904725193977}
        & 27.1
        & \mround{10.161297819} 
        & \mround{0.3077993965089321}
        \\
        \hline
        
        NBB + Direct copying
        & \mround{69.99023846}
        & \mround{0.07978755904480349}
        & 8.9
        & \mround{46.44052433} 
        & \mround{0.08501235633322503}
        \\

        NBB + Laplace blending
        & \mround{69.62766699}
        & \mround{0.08020854576616548}
        & 12.2
        & \mround{31.18230837}
        & \mround{0.08513711040704511}
        \\
        \hline
        
        Unmodified model 
        & \na
        & \na
        & 63.8
        & \na 
        &  \na
        \end{tabular}
}
    }
    \caption{We edit a StyleGANv2 \cite{karras2020styleganv2} LSUN church \cite{yu2015lsun} model to replace domes with spires/trees in $n=10,000$ images. To quantify efficacy, we show the percentage of \texttt{dome} category pixels changed to the target category, determined by a segmenter ~\cite{xiao2018unified}. To quantify undesired changes, we report LPIPS distance between edited and unchanged images, in non-dome regions. For realism, workers make $n=1,000$ pairwise judgements comparing images from other methods to ours.
    }
    \label{tab:church}
\end{table}

\myparagraph{Quantitative Evaluation.} In Tables~\ref{tab:ffhq}~and~\ref{tab:church},%
 we compare the results to several baselines.  We compare our method to the naive approach of fine-tuning all weights according to \refeq{ideal-constraint}, as well as the method of optimizing all the weights of a layer without constraining the direction of the change, as in \refeq{lconstraint}, 
and to a state-of-the-art image alignment algorithm, \emph{Neural Best-Buddies} (NBB~\cite{aberman2018neural}), which is used to propagate an edit across a set of similar images by compositing pixels according to identified sparse correspondences. To transfer an edit from a target image, we use NBB and Moving Least Squares~\cite{schaefer2006moving} to compute a dense correspondence between the source image we would like to edit and the original target image. We use this dense correspondence field to warp the masked target into the source image.
We test both direct copying and Laplace blending. 

For each setting, we measure the efficacy of the edits on a sample of 10, 000 generated images, and we also quantify the undesired changes made by each method. For the smiling edit, we measure efficacy by counting images classified as smiling by an attribute classifier~\cite{sharma2020slim}, and we also quantify changes made in the images outside the mouth region by masking lips using a face segmentation model~\cite{zll2019faceparsing} and using LPIPS~\cite{zhang2018unreasonable} to quantify changes.  For the dome edits, we measure how many dome pixels are judged to be changed to non-domes by a segmentation model~\cite{xiao2018unified}, and we measure undesired changes outside dome areas using LPIPS. We also conduct a user study where users are asked to compare the realism of our edited output to the same image edited using baseline methods. We find that our method produces more realistic outputs that are more narrowly targeted than the baseline methods. For the smile edit, our method is not as aggressive as baseline methods at introducing smiles, but for the dome edits, our method is more effective than baseline methods at executing the change. Our metrics are further discussed in \refapp{imp-details}.

\begin{figure}[t]
    \centering
    \includegraphics[width=0.95\textwidth]{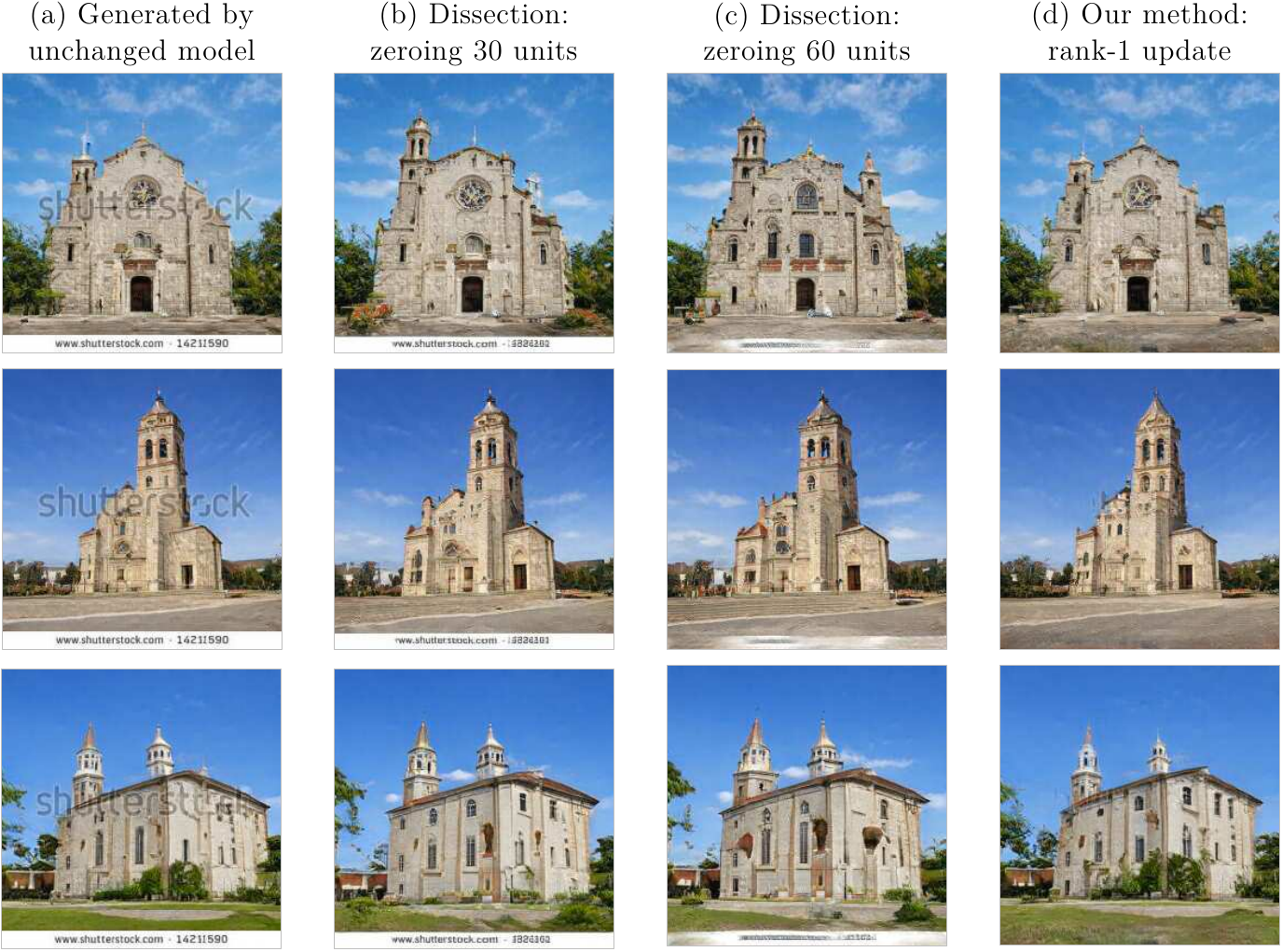}\vspace{-1mm}
    \caption{Removing watermarks from StyleGANv2 \cite{karras2020styleganv2} LSUN church \cite{yu2015lsun} model.  (a) Many images generated by this model include transparent watermarks in the center or text on the bottom. (b) Using GAN Dissection~\cite{bau2019gandissect} to zero 30 text-specific units removes middle but not bottom text cleanly. (c) Removing 60 units does not fully remove text, and distorts other aspects of the image.  (b) Applying our method to create a rank-1 change erases both middle and bottom text cleanly. }
    \label{fig:watermark-removal}%
\end{figure}
\begin{table}[t]
    \centering \small
    \begin{tabular}{l c c}
    Count of visible watermarks   &  middle & bottom  \\ \hline
    Zeroing 30 units (GAN Dissection) & \textbf{0} & 6 \\
    Zeroing 60 units (GAN Dissection) & \textbf{0} & 4 \\
    Rank-1 update (our method) & \textbf{0} & \textbf{0} \\ \hline
    Unmodified model &  64 & 26
    \vspace{1mm}
    \end{tabular}
    
    \caption{Visible watermark text produced by StyleGANv2 church model in $n=1000$ images, without modification, with sets of units zeroed (using the method of GAN Dissection), and using our method to apply a rank-one update.}
    \label{tab:watermarks}
     \vspace{-10mm}
\end{table}
\subsection{Removing undesired features}
\lblsec{exp-watermark}
Here we test our method on the removal of undesired features.  \reffig{watermark-removal}a shows several examples of images output by a pre-trained StyleGANv2 church model. This model occasionally synthesizes images with text overlaid in the middle and the bottom resembling stock-photo watermarks in the training set.

The GAN Dissection study~\cite{bau2019gandissect} has shown that some objects can be removed from a generator by zeroing the units that best match those objects.  To find these units, we annotated the middle and bottom text regions in ten generated images, and we identified a set of 60 units that are most highly correlated with features in these regions.  Zeroing the most correlated 30 units removes some of the text, but leaves much bottom text unremoved, as shown in \reffig{watermark-removal}b.  Zeroing all 60 units reduces more of the bottom text but begins to alter the main content of the images, as shown in \reffig{watermark-removal}c.

For our method, we use the ten user-annotated images as a context to create a rank-one constraint direction $d$ for updating the model, and as an optimization target $K_* \rightarrow V_*$, we use one successfully removed watermark from the setting shown in \reffig{watermark-removal}b.  Since our method applies a narrow rank-1 change constraint, it would be expected to produce a loose approximation of the rank-30 change in the training example.  Yet we find that it has instead improved specificity and generalization of watermark removal, removing both middle and bottom text cleanly while introducing few changes in the main content of the image. We repeat the process for 1000 images and tabulate the results in Table~\ref{tab:watermarks}.

\subsection{Changing contextual rules}
\lblsec{exp-reflection}
\begin{figure}[t]
    \centering
    \includegraphics[width=1.0\textwidth,trim=0 10pt 0 0]{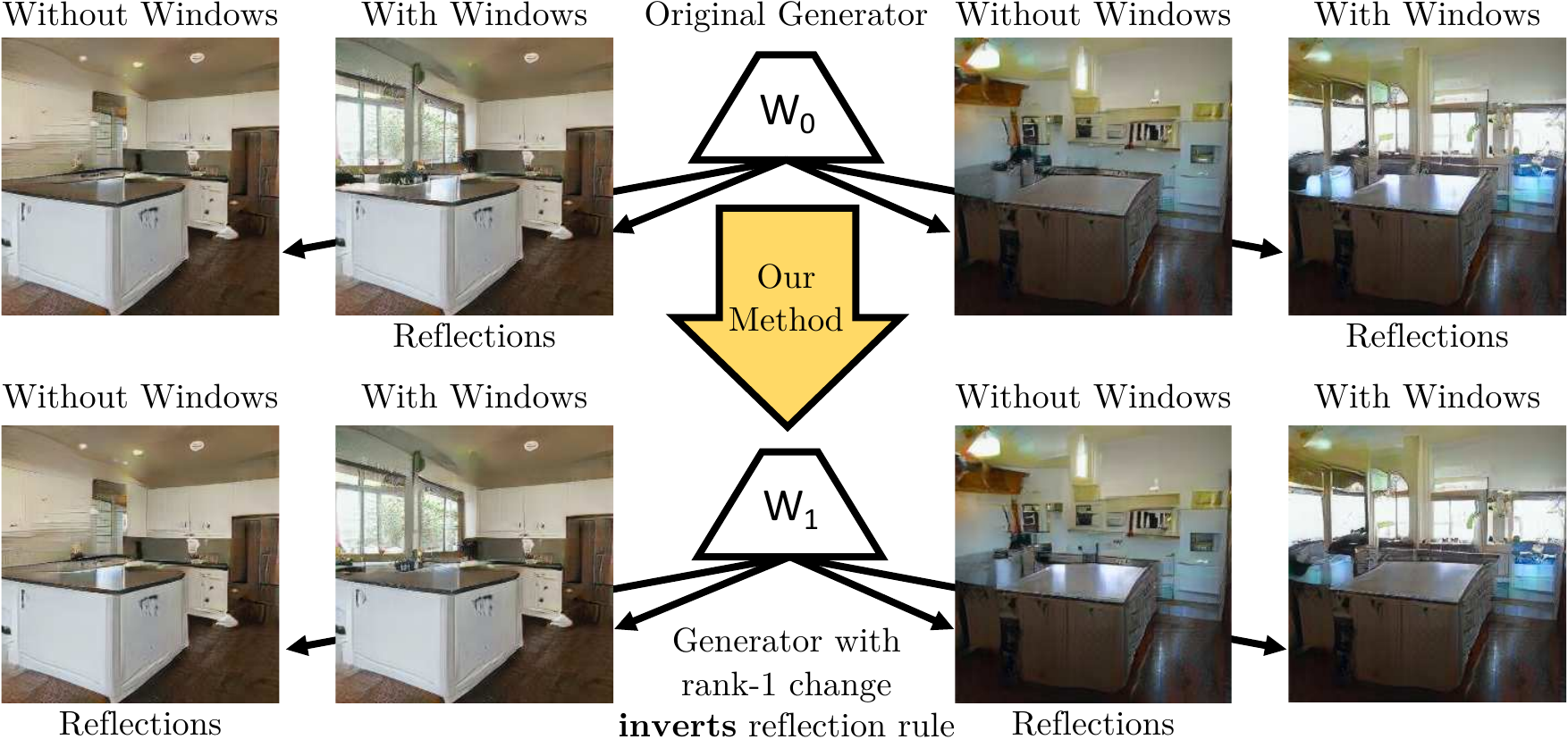}
    \caption{Inverting a single semantic rule within a model.  At the top row, a Progressive GAN~\cite{karras2018progressive} trained on LSUN kitchens~\cite{yu2015lsun} links windows to reflections: when windows are added by manipulating intermediate features identified by GAN Dissection~\cite{bau2019gandissect}, reflections appear on the table.  In the bottom row, one rule has been changed within the model to {\em invert} the relationship between windows and reflections.  Now adding windows \emph{decreases} reflections and vice-versa.}
    \label{fig:reflection-rule}
\end{figure}
In this experiment, we find and alter a rule that determines the illumination interactions between two objects at different locations in an image.

State-of-the-art generative models learn to enforce many relationships between distant objects.  For example, it has been observed~\cite{bau2019ganpaint} that a kitchen-scene Progressive GAN model~\cite{karras2018progressive} enforces a relationship between windows on walls and specular reflections on tables.  When windows are added to a wall, reflections will be added to shiny tabletops, and vice-versa, as illustrated in the first row of \reffig{reflection-rule}.  Thus the model contains a rule that approximates the physical propagation of light in a scene.

In the following experiment, we identified an update direction that allows us to change this model of light reflections.  Instead of specifying an objective that copies an object from one context to another, we used a similar tool to specify a $K_*\rightarrow V_*$ objective that swaps bright tabletop reflections with dim reflections on a set of 15 pairs of scenes that are identical other than the presence or absence of bright windows.  To identify a rank-one change direction $d$, we used projected gradient descent, as described in \refsec{nonlinear}, using SVD to limit the change to rank one during optimization.  The results are shown in the second row of  \reffig{reflection-rule}.  The modified model differs from the original only in a single update direction of a single layer, but it inverts the relationship between windows and reflections: when windows are added, reflections are reduced, and vice-versa.

\section{Discussion}

Machine learning requires data, so how can we create effective models for data that do not yet exist?  Thanks to the rich internal structure of recent GANs, in this paper,  we have found it feasible to create such models by rewriting the rules within existing networks.  Although we may never have seen a tree sprouting from a tower, our network contains rules for both trees and towers, and we can easily create a model that connects those compositional rules to synthesize an endless distribution of images containing the new combination.
 
The development of sophisticated generative models beyond the image domain, such as the GPT-3 language model~\cite{brown2020gpt3} and WaveNet for audio synthesis~\cite{oord2016wavenet}, means that it will be increasingly attractive to rewrite rules within other types of models as well. After training on vast datasets, large-scale deep networks have proven to be capable of representing an extensive range of different styles, sentiments, and topics. Model rewriting provides an avenue for using this structure as a
rich medium for creating
novel kinds of content, behavior, and interaction.

\medskip{\myparagraph{Acknowledgements.} We thank Jonas Wulff, Hendrik Strobelt, Aaron Hertzman, Taesung Park, William Peebles, Gerald Sussman, and William T. Freeman for their vision, encouragement, and many valuable discussions, and we especially thank Piper Bau for video editing. We are grateful for the support of DARPA XAI FA8750-18-C-0004, DARPA SAIL-ON HR0011-20-C-0022, NSF 1524817 on Advancing Visual Recognition with Feature Visualizations, NSF BIGDATA 1447476, and a hardware donation from NVIDIA.}

\clearpage
\bibliographystyle{splncs04}
\bibliography{egbib}

\begin{thebibliography}{10}
\providecommand{\url}[1]{\texttt{#1}}
\providecommand{\urlprefix}{URL }
\providecommand{\doi}[1]{https://doi.org/#1}

\bibitem{aberman2018neural}
Aberman, K., Liao, J., Shi, M., Lischinski, D., Chen, B., Cohen-Or, D.: Neural
  best-buddies: Sparse cross-domain correspondence. ACM Transactions on
  Graphics  \textbf{37}(4), ~69 (2018)

\bibitem{an2008appprop}
An, X., Pellacini, F.: Appprop: all-pairs appearance-space edit propagation.
  ACM Transactions on Graphics  \textbf{27}(3), ~40 (2008)

\bibitem{anderson1972simple}
Anderson, J.A.: A simple neural network generating an interactive memory.
  Mathematical biosciences  \textbf{14}(3-4),  197--220 (1972)

\bibitem{andrychowicz2016learning}
Andrychowicz, M., Denil, M., Gomez, S., Hoffman, M.W., Pfau, D., Schaul, T.,
  Shillingford, B., De~Freitas, N.: Learning to learn by gradient descent by
  gradient descent. In: NeurIPS (2016)

\bibitem{barnes2009patchmatch}
Barnes, C., Shechtman, E., Finkelstein, A., Goldman, D.B.: Patchmatch: A
  randomized correspondence algorithm for structural image editing. ACM
  Transactions on Graphics  \textbf{28}(3), ~24 (2009)

\bibitem{bau2019ganpaint}
Bau, D., Strobelt, H., Peebles, W., Wulff, J., Zhou, B., Zhu, J., Torralba, A.:
  Semantic photo manipulation with a generative image prior. ACM Transactions
  on Graphics  \textbf{38}(4) (2019)

\bibitem{bau2019gandissect}
Bau, D., Zhu, J.Y., Strobelt, H., Bolei, Z., Tenenbaum, J.B., Freeman, W.T.,
  Torralba, A.: Gan dissection: Visualizing and understanding generative
  adversarial networks. In: ICLR (2019)

\bibitem{bengio1992optimization}
Bengio, S., Bengio, Y., Cloutier, J., Gecsei, J.: On the optimization of a
  synaptic learning rule. In: Optimality in Artificial and Biological Neural
  Networks. pp.~6--8. Univ. of Texas (1992)

\bibitem{bousmalis2017unsupervised}
Bousmalis, K., Silberman, N., Dohan, D., Erhan, D., Krishnan, D.: Unsupervised
  pixel-level domain adaptation with generative adversarial networks. In: CVPR
  (2017)

\bibitem{brown2020gpt3}
Brown, T.B., Mann, B., Ryder, N., Subbiah, M., Kaplan, J., Dhariwal, P.,
  Neelakantan, A., Shyam, P., Sastry, G., Askell, A., et~al.: Language models
  are few-shot learners. arXiv preprint arXiv:2005.14165  (2020)

\bibitem{burt1983laplacian}
Burt, P., Adelson, E.: The laplacian pyramid as a compact image code. IEEE
  Transactions on communications  \textbf{31}(4),  532--540 (1983)

\bibitem{chan2019everybody}
Chan, C., Ginosar, S., Zhou, T., Efros, A.A.: Everybody dance now. In: ICCV
  (2019)

\bibitem{chen2014sparse}
Chen, X., Zou, D., Li, J., Cao, X., Zhao, Q., Zhang, H.: Sparse dictionary
  learning for edit propagation of high-resolution images. In: CVPR (2014)

\bibitem{chen2012manifold}
Chen, X., Zou, D., Zhao, Q., Tan, P.: Manifold preserving edit propagation. ACM
  Transactions on Graphics  \textbf{31}(6), ~1--7 (2012)

\bibitem{cohn1994improving}
Cohn, D., Atlas, L., Ladner, R.: Improving generalization with active learning.
  Machine learning  \textbf{15}(2),  201--221 (1994)

\bibitem{collins2020editing}
Collins, E., Bala, R., Price, B., Susstrunk, S.: Editing in style: Uncovering
  the local semantics of gans. In: CVPR (2020)

\bibitem{donahue2014decaf}
Donahue, J., Jia, Y., Vinyals, O., Hoffman, J., Zhang, N., Tzeng, E., Darrell,
  T.: Decaf: A deep convolutional activation feature for generic visual
  recognition. In: ICML (2014)

\bibitem{dosovitskiy2016generating}
Dosovitskiy, A., Brox, T.: Generating images with perceptual similarity metrics
  based on deep networks. In: Neural Information Processing Systems (2016)

\bibitem{dy2000visualization}
Dy, J.G., Brodley, C.E.: Visualization and interactive feature selection for
  unsupervised data. In: SIGKDD (2000)

\bibitem{efros2001image}
Efros, A.A., Freeman, W.T.: Image quilting for texture synthesis and transfer.
  In: ACM SIGGRAPH (2001)

\bibitem{endo2016deepprop}
Endo, Y., Iizuka, S., Kanamori, Y., Mitani, J.: Deepprop: Extracting deep
  features from a single image for edit propagation. Computer Graphics Forum
  \textbf{35}(2),  189--201 (2016)

\bibitem{fails2003interactive}
Fails, J.A., Olsen~Jr, D.R.: Interactive machine learning. In: ACM IUI (2003)

\bibitem{finn2017model}
Finn, C., Abbeel, P., Levine, S.: Model-agnostic meta-learning for fast
  adaptation of deep networks. In: ICML. pp. 1126--1135. JMLR. org (2017)

\bibitem{fried2019text}
Fried, O., Tewari, A., Zollh{\"o}fer, M., Finkelstein, A., Shechtman, E.,
  Goldman, D.B., Genova, K., Jin, Z., Theobalt, C., Agrawala, M.: Text-based
  editing of talking-head video. ACM Transactions on Graphics  \textbf{38}(4),
  1--14 (2019)

\bibitem{gatys2015neural}
Gatys, L.A., Ecker, A.S., Bethge, M.: Image style transfer using convolutional
  neural networks. In: CVPR (2016)

\bibitem{goodfellow2014generative}
Goodfellow, I., Pouget-Abadie, J., Mirza, M., Xu, B., Warde-Farley, D., Ozair,
  S., Courville, A., Bengio, Y.: Generative adversarial nets. In: Neural
  Information Processing Systems (2014)

\bibitem{guo2003coordinating}
Guo, D.: Coordinating computational and visual approaches for interactive
  feature selection and multivariate clustering. Information Visualization
  \textbf{2}(4),  232--246 (2003)

\bibitem{hasinoff2010search}
Hasinoff, S.W., J{\'o}{\'z}wiak, M., Durand, F., Freeman, W.T.:
  Search-and-replace editing for personal photo collections. In: ICCP (2010)

\bibitem{hertzmann2001image}
Hertzmann, A., Jacobs, C.E., Oliver, N., Curless, B., Salesin, D.H.: Image
  analogies. In: ACM SIGGRAPH (2001)

\bibitem{heusel2017gans}
Heusel, M., Ramsauer, H., Unterthiner, T., Nessler, B., Hochreiter, S.: Gans
  trained by a two time-scale update rule converge to a local nash equilibrium.
  In: Neural Information Processing Systems (2017)

\bibitem{Hinton06}
Hinton, G.E., Osindero, S., Teh, Y.W.: A fast learning algorithm for deep
  belief nets. Neural Computation  \textbf{18},  1527--1554 (2006)

\bibitem{huang2018multimodal}
Huang, X., Liu, M.Y., Belongie, S., Kautz, J.: Multimodal unsupervised
  image-to-image translation. ECCV  (2018)

\bibitem{huh2020ganprojection}
Huh, M., Zhang, R., Zhu, J.Y., Paris, S., Hertzmann, A.: Transforming and
  projecting images to class-conditional generative networks. In: ECCV (2020)

\bibitem{iizuka2016let}
Iizuka, S., Simo-Serra, E., Ishikawa, H.: {Let there be Color!: Joint
  End-to-end Learning of Global and Local Image Priors for Automatic Image
  Colorization with Simultaneous Classification}. ACM Transactions on Graphics
  \textbf{35}(4) (2016)

\bibitem{isola2017image}
Isola, P., Zhu, J.Y., Zhou, T., Efros, A.A.: Image-to-image translation with
  conditional adversarial networks. In: CVPR (2017)

\bibitem{jahanian2020steerability}
Jahanian, A., Chai, L., Isola, P.: On the" steerability" of generative
  adversarial networks. In: ICLR (2020)

\bibitem{jiang2017interactive}
Jiang, B., Canny, J.: Interactive machine learning via a gpu-accelerated
  toolkit. In: ACM IUI. pp. 535--546 (2017)

\bibitem{johnson2016perceptual}
Johnson, J., Alahi, A., Fei-Fei, L.: Perceptual losses for real-time style
  transfer and super-resolution. In: ECCV (2016)

\bibitem{kapoor2010interactive}
Kapoor, A., Lee, B., Tan, D., Horvitz, E.: Interactive optimization for
  steering machine classification. In: Proceedings of the SIGCHI Conference on
  Human Factors in Computing Systems. pp. 1343--1352 (2010)

\bibitem{karras2019ffhq}
Karras, T.: {FFHQ} dataset. \url{https://github.com/NVlabs/ffhq-dataset} (2019)

\bibitem{karras2018progressive}
Karras, T., Aila, T., Laine, S., Lehtinen, J.: Progressive growing of gans for
  improved quality, stability, and variation. In: ICLR (2018)

\bibitem{karras2019style}
Karras, T., Laine, S., Aila, T.: A style-based generator architecture for
  generative adversarial networks. In: CVPR (2019)

\bibitem{karras2020styleganv2}
Karras, T., Laine, S., Aittala, M., Hellsten, J., Lehtinen, J., Aila, T.:
  Analyzing and improving the image quality of stylegan. In: CVPR (2020)

\bibitem{kingma2014adam}
Kingma, D., Ba, J.: Adam: A method for stochastic optimization. In: ICLR (2015)

\bibitem{kingma2014auto}
Kingma, D.P., Welling, M.: Auto-encoding variational bayes. ICLR  (2014)

\bibitem{kohonen1972correlation}
Kohonen, T.: Correlation matrix memories. IEEE transactions on computers
  \textbf{100}(4),  353--359 (1972)

\bibitem{kohonen2012associative}
Kohonen, T.: Associative memory: A system-theoretical approach, vol.~17.
  Springer Science \& Business Media (2012)

\bibitem{kohonen1973representation}
Kohonen, T., Ruohonen, M.: Representation of associated data by matrix
  operators. IEEE Transactions on Computers  \textbf{100}(7),  701--702 (1973)

\bibitem{kokiopoulou2011trace}
Kokiopoulou, E., Chen, J., Saad, Y.: Trace optimization and eigenproblems in
  dimension reduction methods. Numerical Linear Algebra with Applications
  \textbf{18}(3),  565--602 (2011)

\bibitem{krause2014infuse}
Krause, J., Perer, A., Bertini, E.: Infuse: interactive feature selection for
  predictive modeling of high dimensional data. IEEE transactions on
  visualization and computer graphics  \textbf{20}(12),  1614--1623 (2014)

\bibitem{lake2015human}
Lake, B.M., Salakhutdinov, R., Tenenbaum, J.B.: Human-level concept learning
  through probabilistic program induction. Science  \textbf{350}(6266),
  1332--1338 (2015)

\bibitem{larsson2016learning}
Larsson, G., Maire, M., Shakhnarovich, G.: Learning representations for
  automatic colorization. In: ECCV (2016)

\bibitem{levin2004colorization}
Levin, A., Lischinski, D., Weiss, Y.: Colorization using optimization. ACM
  Transactions on Graphics  \textbf{23}(3),  689--694 (2004)

\bibitem{liao2017visual}
Liao, J., Yao, Y., Yuan, L., Hua, G., Kang, S.B.: Visual attribute transfer
  through deep image analogy. ACM Transactions on Graphics  \textbf{36}(4),
  1--15 (2017)

\bibitem{liu2017unsupervised}
Liu, M.Y., Breuel, T., Kautz, J.: Unsupervised image-to-image translation
  networks. In: Neural Information Processing Systems (2017)

\bibitem{luan2017deep}
Luan, F., Paris, S., Shechtman, E., Bala, K.: Deep photo style transfer. In:
  CVPR (2017)

\bibitem{mathieu2016deep}
Mathieu, M., Couprie, C., LeCun, Y.: Deep multi-scale video prediction beyond
  mean square error. In: ICLR (2016)

\bibitem{nagano2018pagan}
Nagano, K., Seo, J., Xing, J., Wei, L., Li, Z., Saito, S., Agarwal, A.,
  Fursund, J., Li, H., Roberts, R., et~al.: pagan: real-time avatars using
  dynamic textures. In: ACM SIGGRAPH Asia. p.~258 (2018)

\bibitem{oord2016wavenet}
Oord, A.v.d., Dieleman, S., Zen, H., Simonyan, K., Vinyals, O., Graves, A.,
  Kalchbrenner, N., Senior, A., Kavukcuoglu, K.: Wavenet: A generative model
  for raw audio. arXiv preprint arXiv:1609.03499  (2016)

\bibitem{park2019SPADE}
Park, T., Liu, M.Y., Wang, T.C., Zhu, J.Y.: Semantic image synthesis with
  spatially-adaptive normalization. In: CVPR (2019)

\bibitem{patel2011using}
Patel, K., Drucker, S.M., Fogarty, J., Kapoor, A., Tan, D.S.: Using multiple
  models to understand data. In: IJCAI (2011)

\bibitem{perez2003poisson}
P{\'e}rez, P., Gangnet, M., Blake, A.: Poisson image editing. In: ACM SIGGRAPH.
  pp. 313--318 (2003)

\bibitem{Faceshop}
Portenier, T., Hu, Q., Szab\'{o}, A., Bigdeli, S.A., Favaro, P., Zwicker, M.:
  Faceshop: Deep sketch-based face image editing. ACM Transactions on Graphics
  \textbf{37}(4),  99:1--99:13 (Jul 2018)

\bibitem{raghavan2006active}
Raghavan, H., Madani, O., Jones, R.: Active learning with feedback on features
  and instances. Journal of Machine Learning Research  \textbf{7}(Aug),
  1655--1686 (2006)

\bibitem{reinhard2001color}
Reinhard, E., Adhikhmin, M., Gooch, B., Shirley, P.: Color transfer between
  images. IEEE Computer graphics and applications  \textbf{21}(5),  34--41
  (2001)

\bibitem{saenko2010adapting}
Saenko, K., Kulis, B., Fritz, M., Darrell, T.: Adapting visual category models
  to new domains. In: ECCV (2010)

\bibitem{schaefer2006moving}
Schaefer, S., McPhail, T., Warren, J.: Image deformation using moving least
  squares. ACM Transactions on Graphics  \textbf{25}(3),  533–540 (Jul 2006)

\bibitem{settles2008analysis}
Settles, B., Craven, M.: An analysis of active learning strategies for sequence
  labeling tasks. In: EMNLP (2008)

\bibitem{sharma2020slim}
Sharma, A., Foroosh, H.: Slim-cnn: A light-weight cnn for face attribute
  prediction. In: International Conference on Automatic Face and Gesture
  Recognition (2020)

\bibitem{shocher2018zero}
Shocher, A., Cohen, N., Irani, M.: “zero-shot” super-resolution using deep
  internal learning. In: CVPR (2018)

\bibitem{socher2013zero}
Socher, R., Ganjoo, M., Manning, C.D., Ng, A.: Zero-shot learning through
  cross-modal transfer. In: Neural Information Processing Systems (2013)

\bibitem{suzuki2018spatially}
Suzuki, R., Koyama, M., Miyato, T., Yonetsuji, T., Zhu, H.: Spatially
  controllable image synthesis with internal representation collaging. arXiv
  preprint arXiv:1811.10153  (2018)

\bibitem{taigman2017unsupervised}
Taigman, Y., Polyak, A., Wolf, L.: Unsupervised cross-domain image generation.
  In: ICLR (2017)

\bibitem{ulyanov2018deep}
Ulyanov, D., Vedaldi, A., Lempitsky, V.: Deep image prior. In: CVPR (2018)

\bibitem{wolberg1990digital}
Wolberg, G.: Digital image warping. IEEE computer society press (1990)

\bibitem{xiao2018unified}
Xiao, T., Liu, Y., Zhou, B., Jiang, Y., Sun, J.: Unified perceptual parsing for
  scene understanding. In: ECCV (2018)

\bibitem{xu2009efficient}
Xu, K., Li, Y., Ju, T., Hu, S.M., Liu, T.Q.: Efficient affinity-based edit
  propagation using kd tree. ACM Transactions on Graphics  \textbf{28}(5),
  ~1--6 (2009)

\bibitem{yosinski2014transferable}
Yosinski, J., Clune, J., Bengio, Y., Lipson, H.: How transferable are features
  in deep neural networks? In: Neural Information Processing Systems (2014)

\bibitem{yu2018bisenet}
Yu, C., Wang, J., Peng, C., Gao, C., Yu, G., Sang, N.: Bisenet: Bilateral
  segmentation network for real-time semantic segmentation. In: ECCV (2018)

\bibitem{yu2015lsun}
Yu, F., Seff, A., Zhang, Y., Song, S., Funkhouser, T., Xiao, J.: Lsun:
  Construction of a large-scale image dataset using deep learning with humans
  in the loop. arXiv preprint arXiv:1506.03365  (2015)

\bibitem{yucer2012transfusive}
Y{\"u}cer, K., Jacobson, A., Hornung, A., Sorkine, O.: Transfusive image
  manipulation. ACM Transactions on Graphics  \textbf{31}(6), ~1--9 (2012)

\bibitem{zeiler2014visualizing}
Zeiler, M.D., Fergus, R.: Visualizing and understanding convolutional networks.
  In: ECCV (2014)

\bibitem{zhang2016colorful}
Zhang, R., Isola, P., Efros, A.A.: Colorful image colorization. In: ECCV (2016)

\bibitem{zhang2018unreasonable}
Zhang, R., Isola, P., Efros, A.A., Shechtman, E., Wang, O.: The unreasonable
  effectiveness of deep features as a perceptual metric. In: CVPR (2018)

\bibitem{zhang2017real}
Zhang, R., Zhu, J.Y., Isola, P., Geng, X., Lin, A.S., Yu, T., Efros, A.A.:
  Real-time user-guided image colorization with learned deep priors. ACM
  Transactions on Graphics  \textbf{9}(4) (2017)

\bibitem{zhu2017unpaired}
Zhu, J.Y., Park, T., Isola, P., Efros, A.A.: Unpaired image-to-image
  translation using cycle-consistent adversarial networks. In: ICCV (2017)

\bibitem{zll2019faceparsing}
ZLL: Face-parsing pytorch.
  \url{https://github.com/zllrunning/face-parsing.PyTorch} (2019)

\end{thebibliography}

\appendix

\noindent\textbf{\LARGE Appendix}\par\bigskip

\section{Additional Editing Examples}
\lblsec{more-examples}

\begin{figure}
    \centering
    \includegraphics[width=\textwidth]{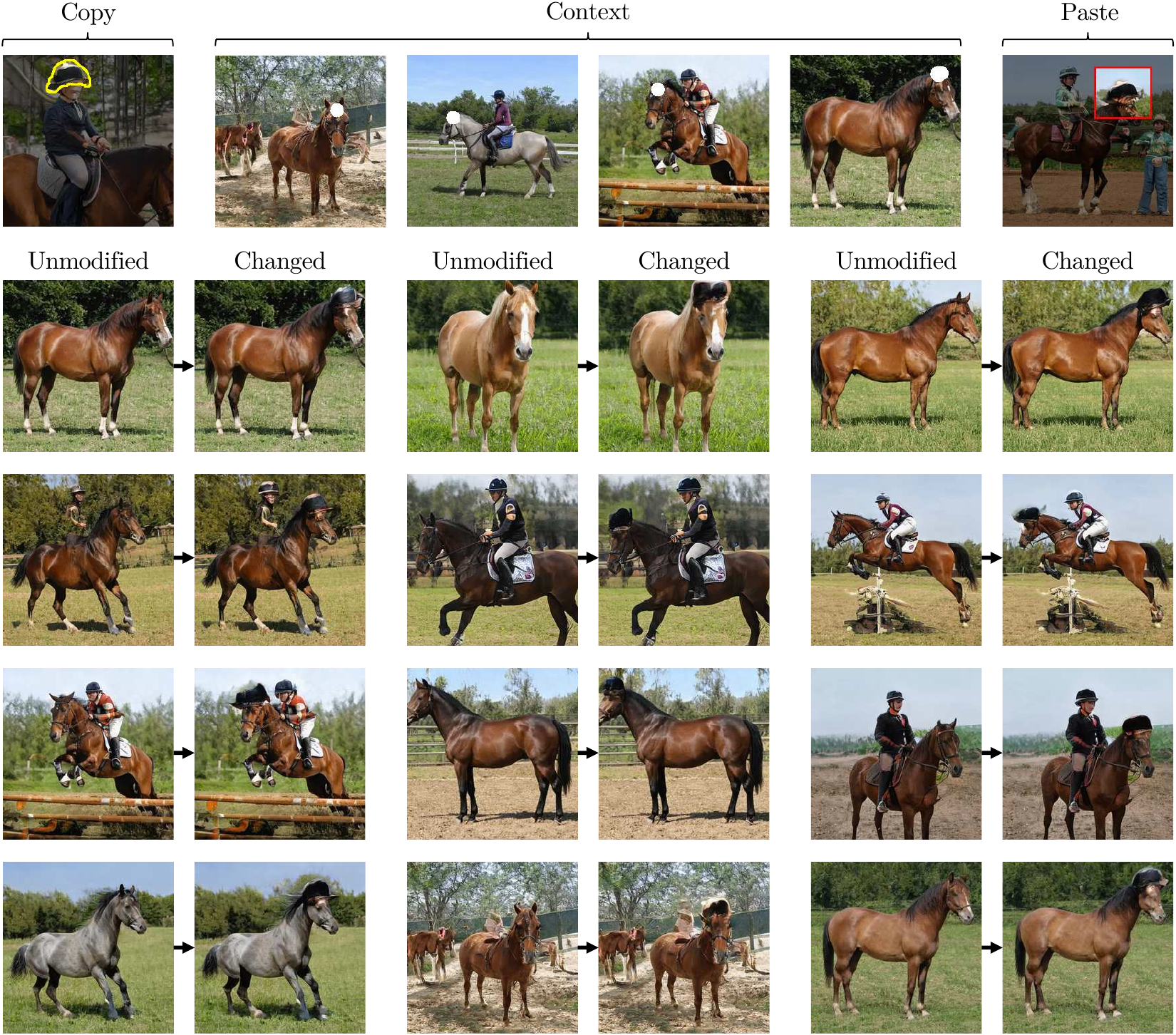}
    \caption{{\bf Giving horses a hat to wear.}  After one hat is pasted onto an example of a horse, and after the user has pointed at four other horse heads, the model is changed so that horses in a variety of poses, settings, shapes, and sizes all get a hat on their head.  This is not merely a re-balancing of the distribution of the model.  This change introduces a new kind of image that was not generated before. The original training data does not include hats on horses, and the original pretrained StyleGANv2 does not synthesize hats on any horses.}
    \label{fig:supp-edit-hat-on-horse}
\end{figure}

\begin{figure}
    \centering
    \includegraphics[width=\textwidth]{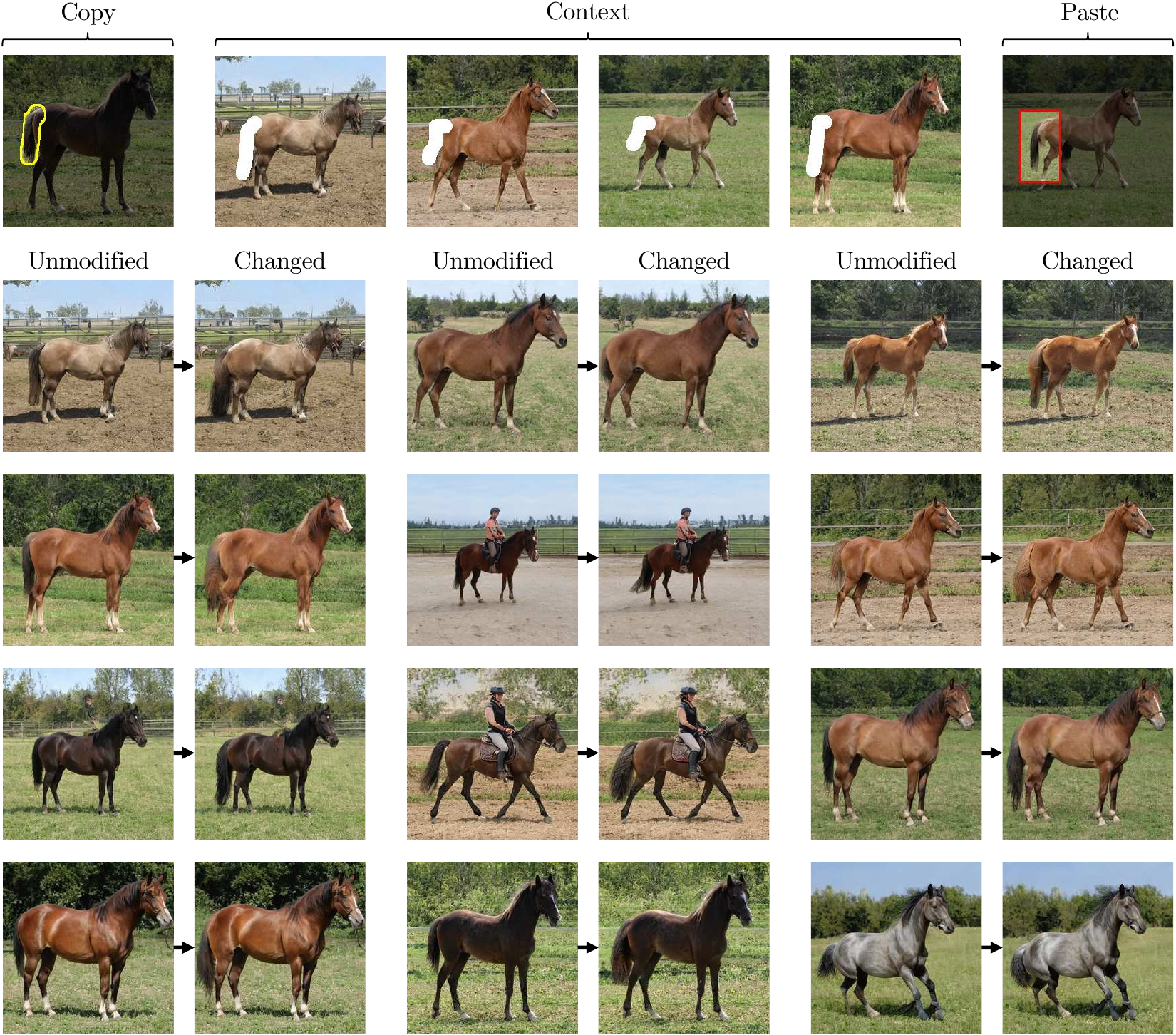}
    \caption{{\bf Giving horses a longer tail.}  Notice that the color, shape, and occlusions of the tail vary to fit the specific horse, but in each case the tail is made longer, as demonstrated in the pasted example.}
    \label{fig:supp-edit-long-horse-tail}
\end{figure}

\begin{figure}
    \centering
    \includegraphics[width=\textwidth]{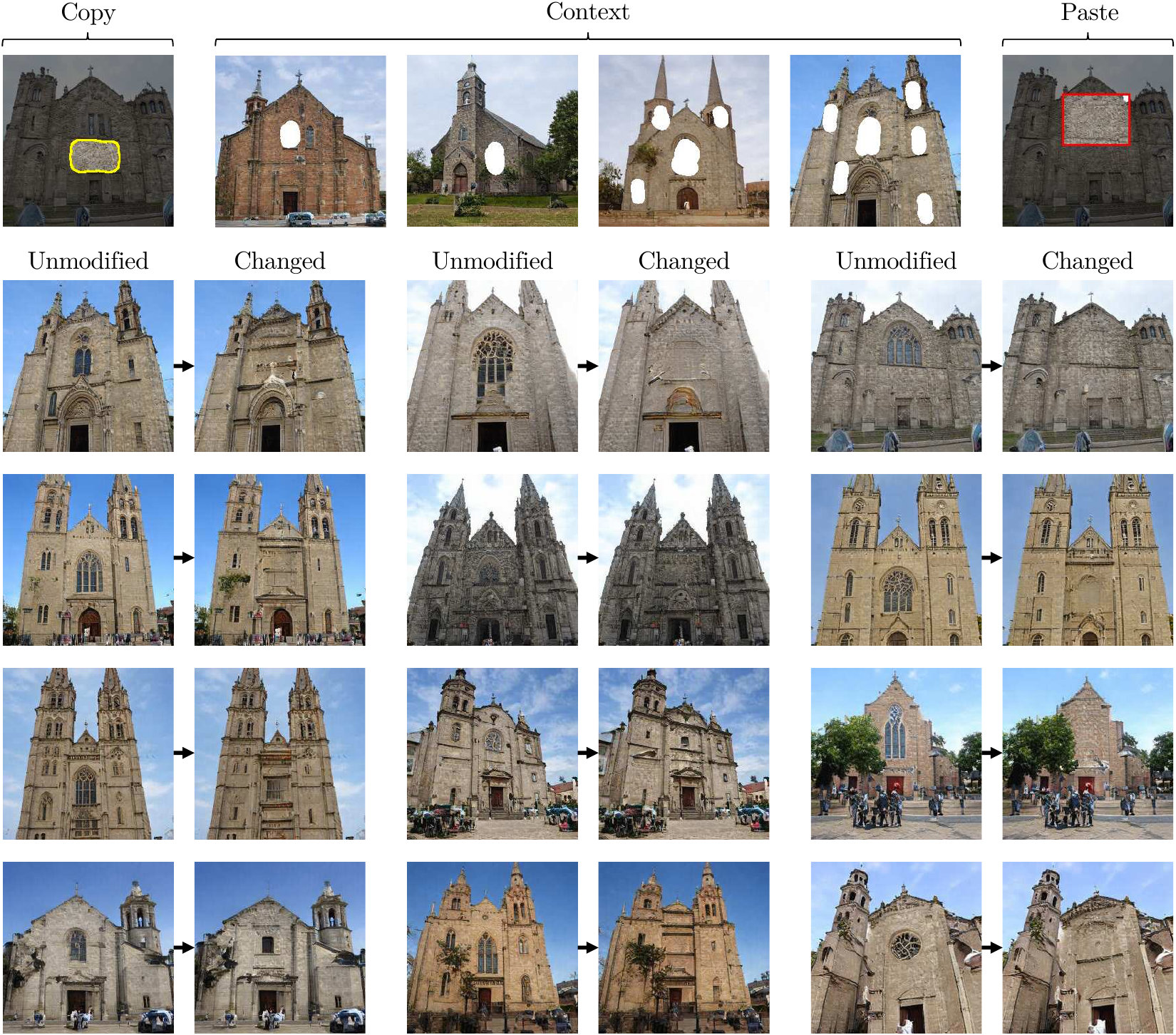}
    \caption{{\bf Removing main windows from churches.}  The modified model will replace the central window with a blank wall, or with a wall with some different details.}
    \label{fig:supp-edit-remove-church-window}
\end{figure}

\begin{figure}
    \centering
    \includegraphics[width=\textwidth]{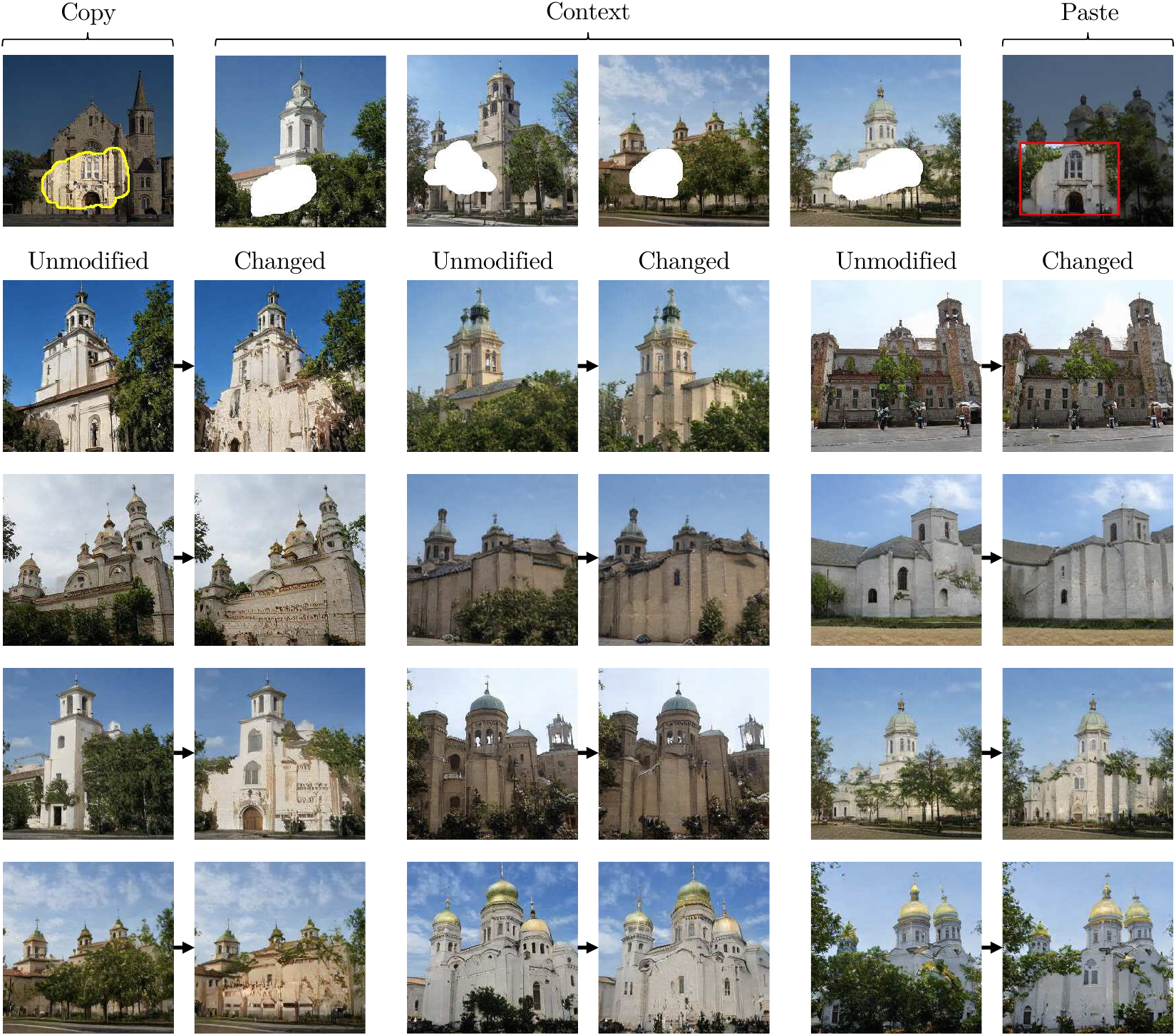}
    \caption{{\bf Reducing the occlusion of buildings by trees.} This edit removes the trees in front of buildings. Note that the model can still synthesize trees next to  buildings. }
    \label{fig:supp-edit-tree-to-building}
\end{figure}

\begin{figure}
    \centering
    \includegraphics[width=\textwidth]{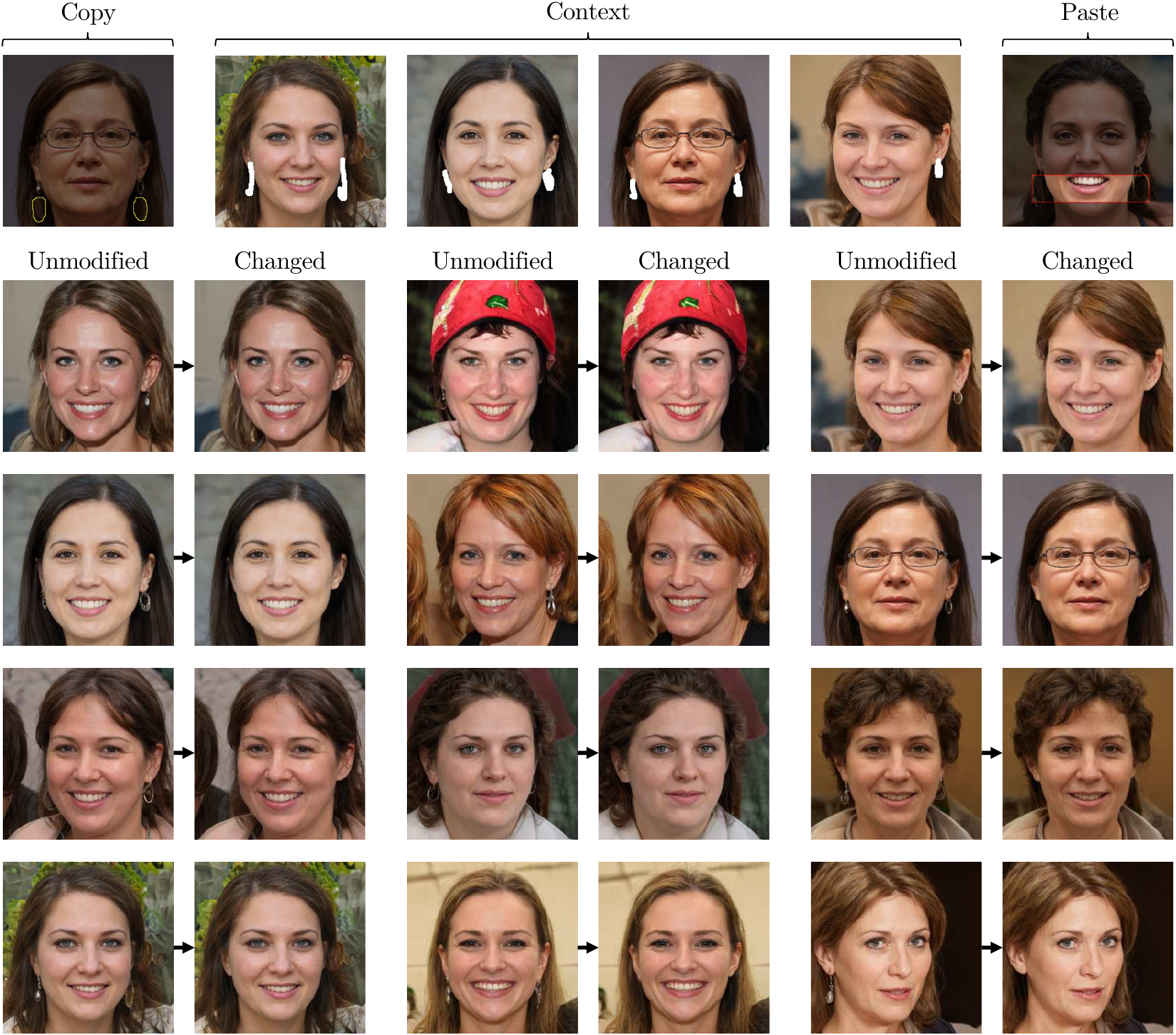}
    \caption{{\bf Removing earrings.}  Removing one set of earrings generalizes to many different types of earrings appearing in different poses.}
    \label{fig:supp-edit-remove-earrings}
\end{figure}

\begin{figure}
    \centering
    \includegraphics[width=\textwidth]{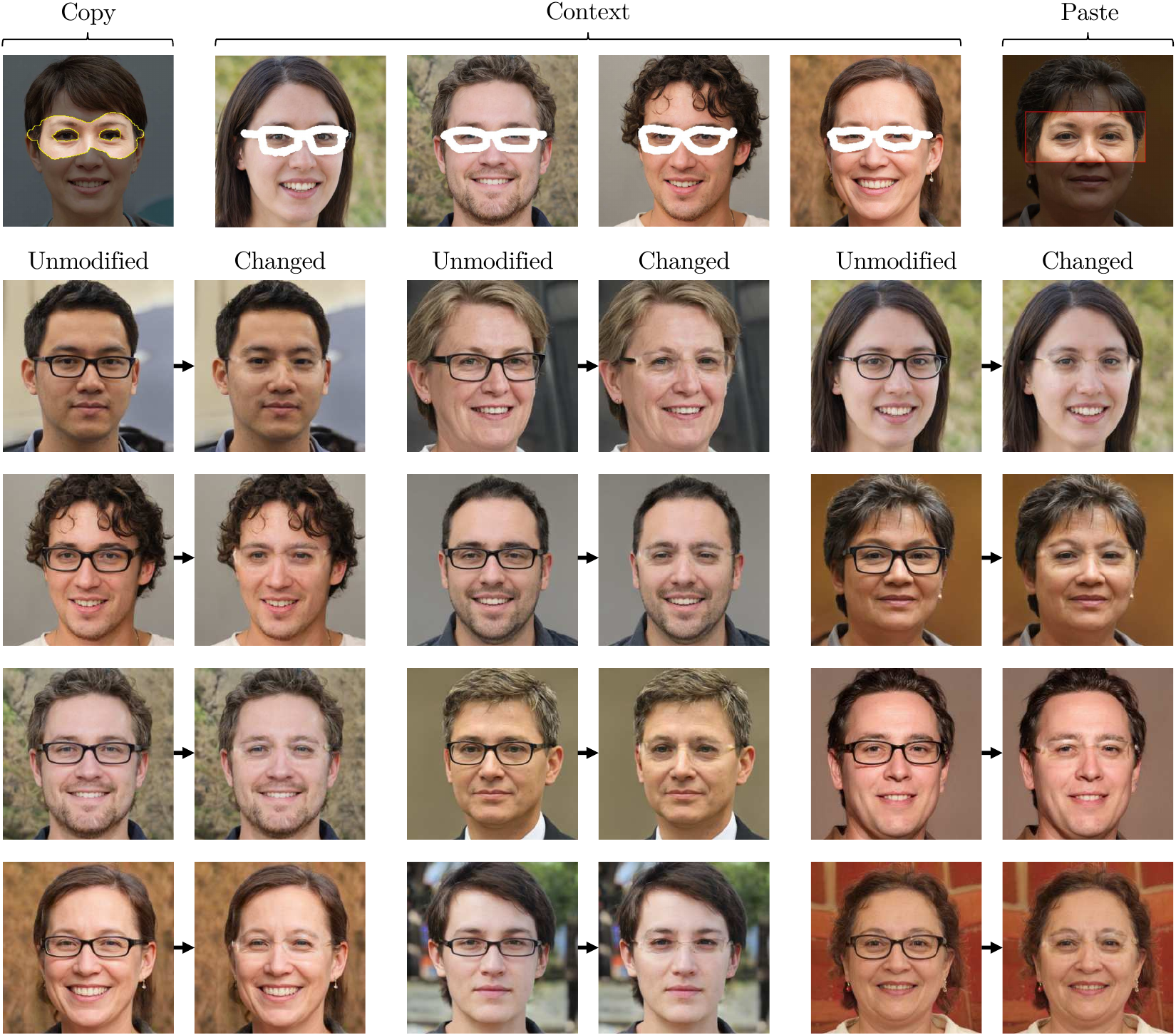}
    \caption{{\bf Removing glasses.}  Note that glasses of different shapes are removed, and most facial structure is recovered.  This is a rank-three change.  Although most of the glasses have been removed,  this edit did not remove the temples (side parts) of some glasses, and did not remove refraction effects.}
    \label{fig:supp-edit-remove-glasses}
\end{figure}

Figures~\ref{fig:supp-edit-hat-on-horse},~\ref{fig:supp-edit-long-horse-tail},~\ref{fig:supp-edit-remove-church-window},~\ref{fig:supp-edit-tree-to-building},~\ref{fig:supp-edit-remove-earrings}, and~\ref{fig:supp-edit-remove-glasses} show additional results of our editing method to change a model to achieve a variety of effects across an entire distribution of generated images. Each figure illustrates a single low-rank change of a StyleGAN v2 model derived from the user gestures shown in the top row.  The twelve pairs of images shown below the top row of each figure are the images that score highest in the context direction $d$, out of a random sample of 1000: that is, these are images that are most relevant to the user's context selection. For each image, both the output of the unmodified original model and the modified model are shown.  All changes are rank-one changes to the model, except \reffig{supp-edit-tree-to-building}, which is rank ten, and \reffig{supp-edit-remove-glasses}, which is rank three.


\section{Solving for $\Lambda$ Algebraically}
\lblsec{solving-lambda}
To strengthen our intuition, here we describe the closed-form solution for $\Lambda$ in the linear case.  Recall from Equations~\ref{eq:cls-const} and~\ref{eq:rank1-update}:
\begin{align}
W_1 k_* &= v_* \\
W_1 & = W_0 + \Lambda d^{T} 
\end{align}
In the above we have written $d=C^{-1}k_*$ as in~\refeq{d-def} for berevity.  Then we can solve for both $W_1$ and $\Lambda$ simultaneously by rewriting the above system as the following matrix product in block form:
\begin{align}
\left[\begin{array}{@{}c|c@{}}
\\
\quad W_1 \quad\, & \Lambda \\
\,
\end{array}\right]
\left[\begin{array}{@{}c|c@{}}
\\
\quad\; I \quad\;\; & k_* \\
\\ \hline
-d^T \rule{0pt}{2.2ex} & 0
\end{array}\right]
& =
\left[\begin{array}{@{}c|c@{}}
\mathstrut \\
\quad W_0 \quad\, & v_* \\
\mathstrut 
\end{array}\right] \\
\left[\begin{array}{@{}c|c@{}}
\mathstrut \\
\quad W_1 \quad\, & \Lambda \\
\mathstrut 
\end{array}\right]
& =
\left[\begin{array}{@{}c|c@{}}
\mathstrut \\
\quad W_0 \quad\, & v_* \\
\mathstrut 
\end{array}\right]
\left[\begin{array}{@{}c|c@{}}
\\
\quad\; I \quad\;\; & k_* \\
\\ \hline
-d^T \rule{0pt}{2.2ex} & 0
\end{array}\right]^{-1}
\end{align}
In practice, we do not solve this linear system because a neural network layer is nonlinear.  In the nonlinear case, instead of using matrix inversion, $\Lambda$ is found using the optimization in Equation~\ref{eq:nl-ins-rule}.

\section{Implementation details}
\lblsec{imp-details}

\myparagraph{Datasets} To compare identical model edits in different settings, we prepare a small set of saved editing sessions for executing an change.  Each session corresponds to a set of masks that a user has drawn in order to specify a region to copy and paste, together with any number of context regions within generated images for a model.  Benchmark editing sessions are included with the source code.

Large-scale datasets are used only for pretraining the generative models.  The generative models we use are trained on the following datasets.  The face model is trained on Flickr-Faces-HQ (FFHQ) \cite{karras2019ffhq}, a dataset of 70{,}000 1024$\times$1024 face images.  The outdoor church, horse, and kitchen models are trained on LSUN image datasets~\cite{yu2015lsun}.  LSUN provides 126{,}000 church images, 2.2 million kitchen images, and 2 million horse images at resolutions of 256$\times$256 and higher.

\myparagraph{Generators} We rewrite two different generative model architectures: Progressive GAN and StyleGAN v2.  The Progressive GAN generator has 18.3 million parameters and 15 convolutional layers; we edit a model pretrained on LSUN kitchens.  We also edit StyleGAN v2~\cite{karras2020styleganv2}.  StyleGAN v2 has 30 million parameters and 14 convolutional layers (17 layers for the higher-resolution faces model). We edit StyleGAN v2 models trained on FFHQ faces, LSUN churches, and LSUN horses.  All the model weights were those published by the original GAN model authors.  For StyleGAN v2, we apply the truncation trick with multiplier 0.5 when running the model.

\myparagraph{Metrics} To quantify undesired perceptual differences made in edits, we use the Learned Perceptual Image Patch Similarity (LPIPS)~\cite{zhang2018unreasonable} metric to compare unedited images to edited images.  We use the default Alexnet-based LPIPS network weights as published by the original LPIPS authors.  To focus the measurement on undesired changes, we follow the method of the GAN projection work~\cite{huh2020ganprojection} and mask out portions of the image that we intend to change, as identified by a semantic segmentation network.  For faces, we segment the image using a modified BiSeNet~\cite{yu2018bisenet} as published by ZLL as faceparsing-Pytorch~\cite{zll2019faceparsing}.  For churches, we segment the image using the Unified Perceptual Parsing network~\cite{xiao2018unified}.

To quantify the efficacy of the change, we also use pretrained networks.  To detect whether a face image is similing, we use a Slim-CNN~\cite{sharma2020slim} facial attribute classifier.  To determine if domes have successfully been edited to other types of objects, we again use the Unified Perceptual Parsing network, and we count pixels that have changed from being classified as domes to buildings or trees.

\myparagraph{User studies} Human realism measurements are done using Amazon Mechanical Turk (AMT). For each baseline editing method, 500 pairs of images are generated comparing an edited image using our approach to the same image edited using a baseline method, and two AMT workers are asked to judge which of the pair is more realistic, for a total of 1000 comparative judgements.  We do not test the fantastical domes-to-trees edit, which is intended to be unrealistic.

\section{Rank Reduction for $D_S$}
\lblsec{rank-reduction}
In this section we discuss the problem of transforming a user's \emph{context} selection $K \in \R^{N\times T}$ (\refsec{ui}) into a constraint subspace $D_S \in \R^{N\times S}$, where the desired dimensionality $s \ll t$ is smaller than the number of given feature samples $T$ provided in $K$.

We shall think of this as a lossy compression problem.  Use $P$ to denote the probability distribution of the layer $L-1$ features (unconditioned on any user selection), and think of $K$ as a discrete distribution over the user's $t$ context examples.  We can then use cross-entropy $H(K, P)$ to quantify the information in $K$, measured as the message length in a code optimized for the distribution $P$.  To express this information measure in the setting used in \refsec{method}, we will model $P$ as a zero-centered Gaussian distribution $P(k) = (2\pi)^{-n/2} \exp -k^T C^{-1} k/2$ with covariance $C$.

If we the normalize the basis using the ZCA whitening transform $Z$, we can express $P$ as a spherical unit normal distribution in the variable $k' = Zk$.  This yields a concise matrix trace expression for cross entropy:
\begin{align}
\text{Let } C &= U \Sigma U^T \text{ be the eigenvector decomposition} \\
Z & \triangleq C^{-1/2} = U \Sigma^{-1/2} U^T \\
k' & \triangleq Z k \\
K' & \triangleq Z K \\
P(k') & = (2\pi)^{-n/2} \exp( -k'^T k'/2)
\end{align}
\begin{align}
H(K', P) & = \sum_{k' \in K'} \frac{1}{t} \log P(k') \\
& = \frac{1}{2t} \sum_{k' \in K'} k'^T k' + \frac{n}{2t} \log 2\pi\\
& = \frac{1}{2t} \trace\left(K'^T K'\right) + \frac{n}{2t}\log 2\pi
\end{align}
In other words, by assuming a Gaussian model, the information in the user's context selection can be quantified the trace of a symmetric matrix given by inner products over the whitened context selection.

To reduce the rank of the user's context selection, we wish to project the elements of $K'$ by discarding information along the $R=N-S$ most uninformative directions.  Therefore, we seek a matrix $Q^*_R \in \R^{N\times R}$ that has $R$ orthonormal columns, chosen so that the projection of the samples $Q_R^{\,}Q_R^T K'$ minimize cross-entropy with $P$:
\begin{align}
Q_R^* & = \argmin_{Q_R} H(Q_R^{\,}Q_R^T K', P) \\
& = \argmin_{Q_R} \trace\left(K'^T Q_R^{\,}Q_R^T Q_R^{\,}Q_R^T K'\right) \\
& = \argmin_{Q_R} \trace\left(Q_R^T K' K'^T Q_R^{\,} \right)
\lbleq{tracemin}
\end{align}
The trace minimization \refeq{tracemin} is an instance of the well-studied trace optimization problem~\cite{kokiopoulou2011trace} that arises in many dimension-reduction settings. It can be solved by setting the columns of $Q_R^*$ to a basis spanning the space of the eigenvectors for the smallest $R$ eigenvalues of $K'_\textsf{ctx} K'^T$.

Denote by $Q_S^* \in \R^{N\times S}$ the matrix of orthonormal eigenvectors for the $S$ \emph{largest} eigenvalues of $K'_\textsf{ctx} K'^T$.  Then we have $(I - Q_R^*Q_R^{*T})k' = Q_S^*Q_S^{*T} k'$, i.e., erasing the uninteresting directions of $Q_R^*$ is the same as preserving the directions $Q_S^*$.  This is the $S$-dimensional subspace that we seek: it is the maximally informative low-dimensional subspace that captures the user's context selection.

Once we have $Q_S^*$ within the whitened basis, the same subspace can be expressed in unwhitened row space coordinates as:
\begin{align}
D_S = Z^T Q_S^* = Z Q_S^*
\end{align}

\section{Axis-aligned rank reduction for $D_S$}
The identification of axis-aligned units most relevant to a user's context selection can also be analyzed using the same rank-reduction objective as \refsec{rank-reduction}, but with a different family for $P$.  Instead of modeling $P$ as a Gaussian with generic covariance $C$, we now model it as an axis-aligned Gaussian with diagonal covariance $\Sigma = \diag(\sigma_i)$.  Then the optimal basis $Q_S^*$ becomes the unit vectors for the unit directions $e_i$ that maximize the expected ratio
\begin{align}
\sum_{k \in K_\textsf{ctx}} \frac{{(e_i^T k)}^2}{\sigma_i^2}
\end{align}
In \refsec{exp-watermark} this scoring is used to identify the units most relevant to watermarks in order to apply GAN dissection unit ablation.

\section{Experiment Details and Results}

Table~\ref{tab:church} shows the quantitative results of comparing our method with various baselines on editing a StyleGANv2 \cite{karras2020styleganv2} LSUN church \cite{yu2015lsun} model. For both edits, our method modifies the 7th convolution layer of the generator, with Adam optimizer \cite{kingma2014adam}, 0.05 learning rate, $2001$ gradient iterations, and projecting to a low-rank change every $10$ iterations (and also after the optimization loop). For $\texttt{domes}\rightarrow\texttt{trees}$, a rank 1 edit is performed.  (These settings are also the defaults provided in the user interface, and were used for video demos.)  For $\texttt{domes}\rightarrow\texttt{spires}$, a rank $10$ edit is performed.

For the StyleGANv2 FFHQ \cite{karras2019style} edit shown in main paper~\ref{tab:ffhq}, our method modifies the 9th convolution layer of the generator, also with Adam optimizer \cite{kingma2014adam}, 0.05 learning rate, $2001$ gradient iterations, and projecting to a low-rank change every $10$ iterations (and also after the optimization loop).

For all experiments, the baseline that finetunes all weights uses the Adam optimizer \cite{kingma2014adam} with $2001$ iterations and  a learning rate of $10^{-4}$.

\section{Reflection Experiment Details}

In \refsec{exp-reflection}, we found  the rank-one rule reversal change for the abstract window lighting rule as follows.

\begin{enumerate}
\item \textbf{Generation}: we use the GAN to generate 15 images in two ways, one adding windows, and one removing windows, by activating and deactivating window-correlated units.  The window correlated units are identified using dissection~\cite{bau2019gandissect}.
\item \textbf{Annotation}: a user masks illuminated
regions of the 15 images far from the windows that show reflected light that differs between the pairs.
\item \textbf{Optimization}: we optimize a change in the weights of the layer to reverse the behavior of the reflected light in the masked areas, to match dark output when there is a window and bright output when there is no window. This optimization is constrained to one direction by using an SVD reduction to rank one every 10 iterations.
\end{enumerate}

The optimization is computed at each individual layer, and we use the layer that achieves the lowest loss with a rank-one change: for this experiment, this is layer 6 of the model.

\section{Selecting a Layer for Editing}
There are two ways to view a convolutional layer: either as a computation in which information from neighboring locations is combined to detect or produce edges, textures, or shapes; or as a memory in which many independent feature mappings are memorized.

In our paper we have adopted the simple view that a layer acts as an associative memory that maps from one layer's local feature vectors to local patches of feature vectors in the next layer.  This view is appropriate when layer representations have features in which neighboring locations are disentangled from one another. In practice, we find that both ProgressiveGAN and StyleGAN representations have this property.  For example, if a feature patch is rendered in isolation from neighboring features, the network will usually render the same object as it does in the context of the full featuremap.

\begin{figure}
    \centering
    \includegraphics[width=\textwidth]{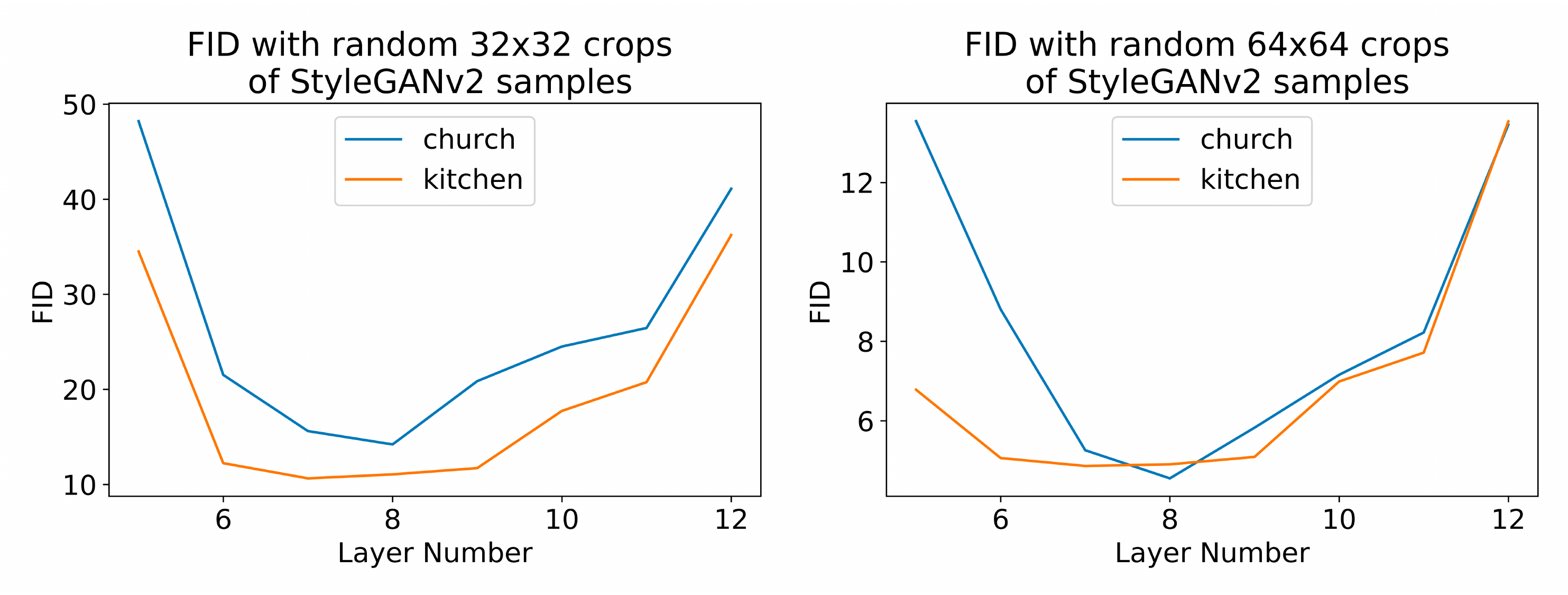}
    \caption{FID of rendered cropped activations with respect to random crops of StyleGANv2 generated images. In StyleGANv2, the $n$th convolutional layer outputs the $n+1$th featuremap layer. The layer numbers above correspond to featuremap layers.}
    \label{fig:supp-stylegan-patch-graph}
\end{figure}

\begin{figure}
    \centering
    \includegraphics[width=\textwidth]{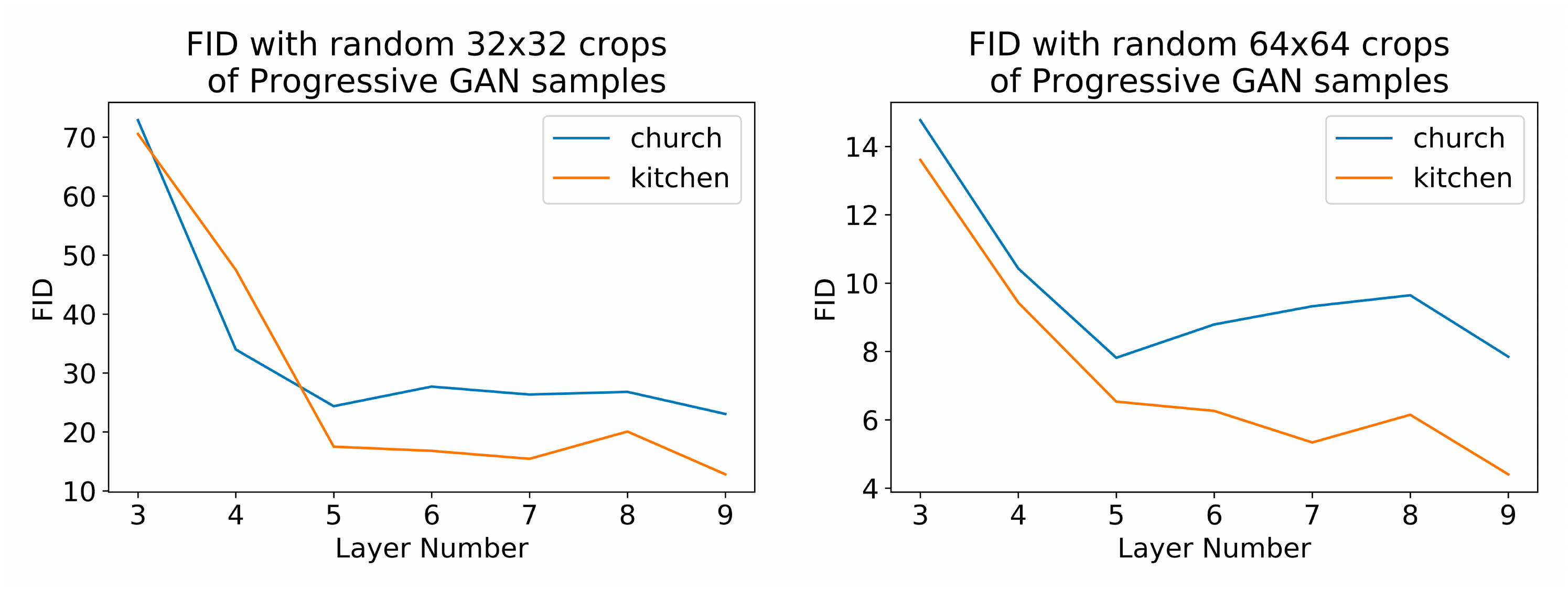}
    \caption{FID of rendered cropped activations with respect to random crops of Progressive GAN generated images.}
    \label{fig:supp-proggan-patch-graph}
\end{figure}

In Figures~\ref{fig:supp-stylegan-patch-graph} and \ref{fig:supp-proggan-patch-graph}, we measure the similarity between patches rendered in isolation compared to same-sized patches cropped out of the full model, using Fr\'echet Inception Distance (FID)~\cite{heusel2017gans}.  Lower FIDs indicate less dependence between neighboring patches, and higher FIDs indicate higher dependence between neighbors.  These graphs show that layers 6-11 in StyleGANv2 and layers 4 and higher in Progressive GAN are most appropriate for editing as an associative memory.  (Note that in StyleGANv2, the $n$th featuremap layer is the output of the $n-1$th convolutional layer, because the first featuremap layer is fixed.  In Progressive GAN, the $n$th featuremap layer is the output of the $n$th convolutional layer.)

\begin{figure}
    \centering
    \includegraphics[width=\textwidth]{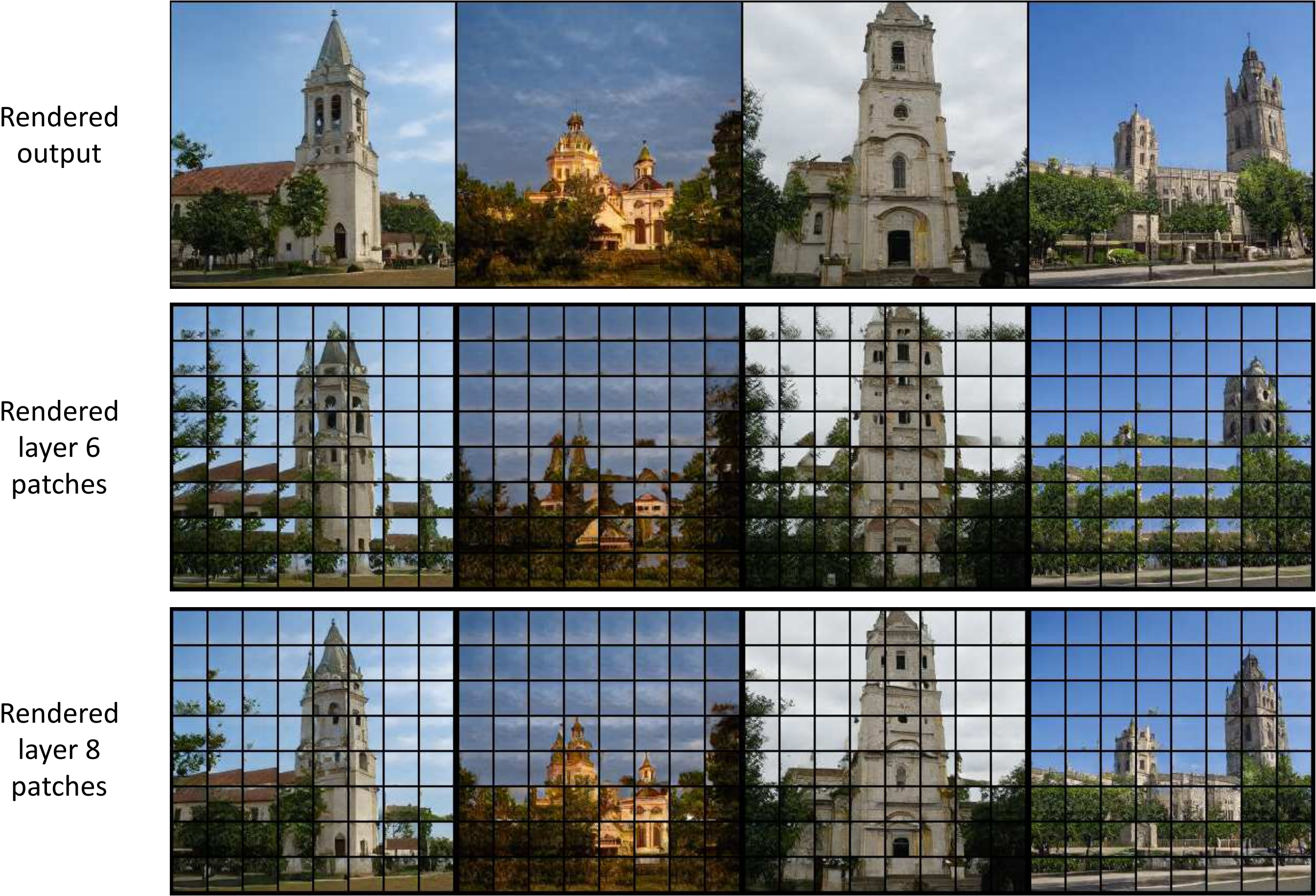}
    \caption{Comparison of rendered cropped activations at various layers of StyleGANv2 generated LSUN church images.}
    \label{fig:supp-stylegan-church-patch}
\end{figure}

\begin{figure}
    \centering
    \includegraphics[width=\textwidth]{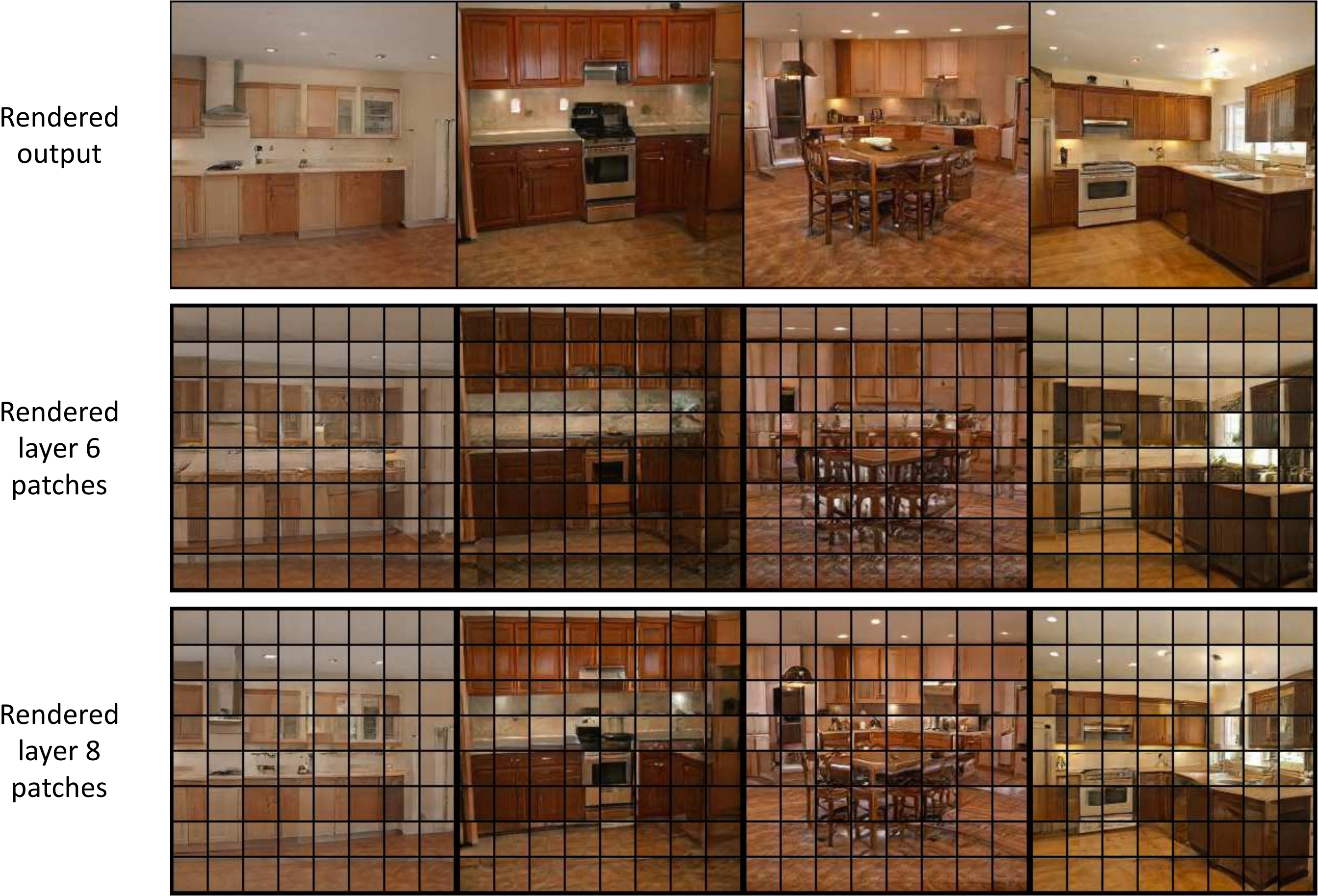}
    \caption{Comparison of rendered cropped activations at various layers of StyleGANv2 generated LSUN kitchen images.}
    \label{fig:supp-stylegan-kitchen-patch}
\end{figure}

\begin{figure}
    \centering
    \includegraphics[width=\textwidth]{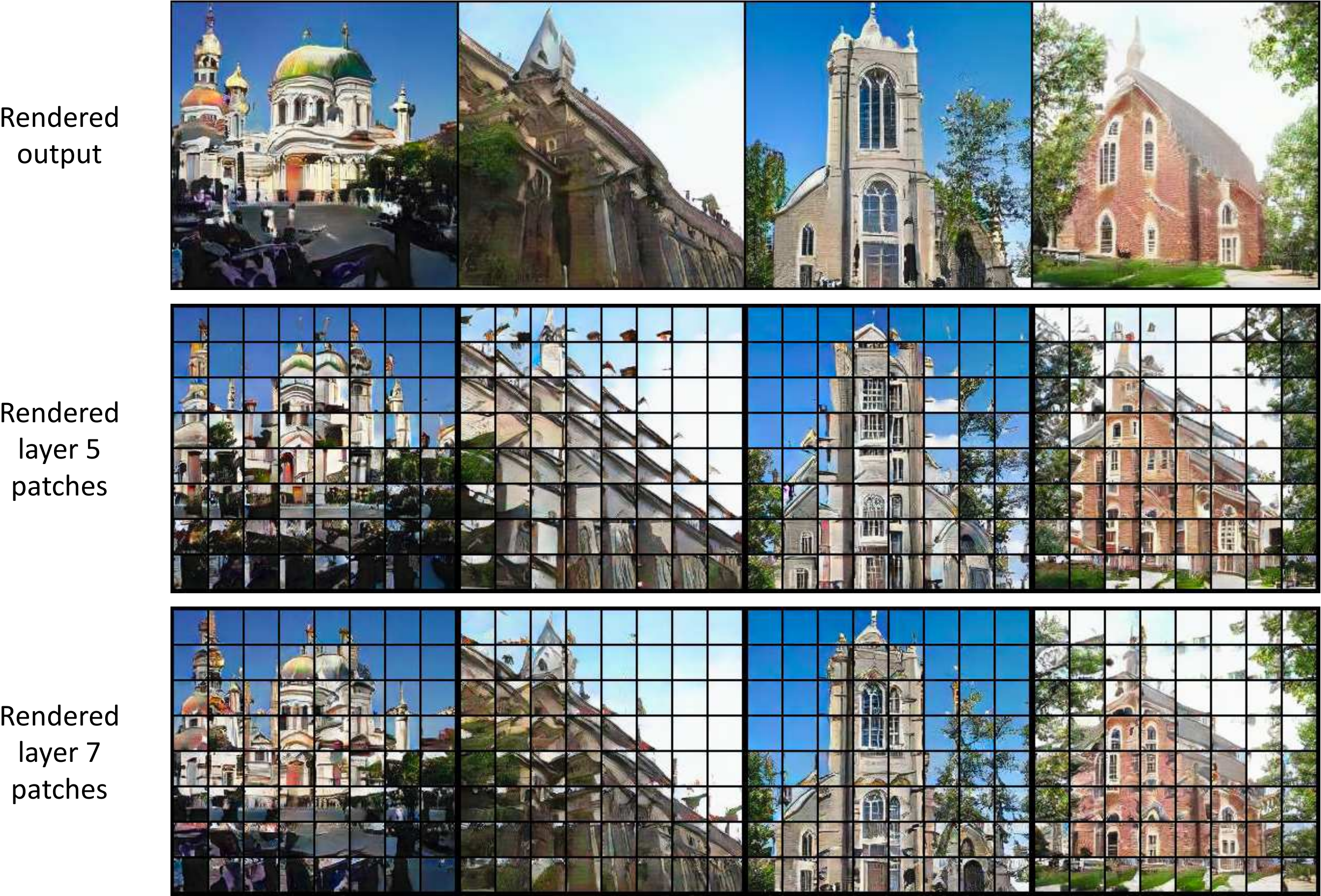}
    \caption{Comparison of rendered cropped activations at various layers of Progressive GAN generated LSUN church images.}
    \label{fig:supp-proggan-church-patch}
\end{figure}

\begin{figure}
    \centering
    \includegraphics[width=\textwidth]{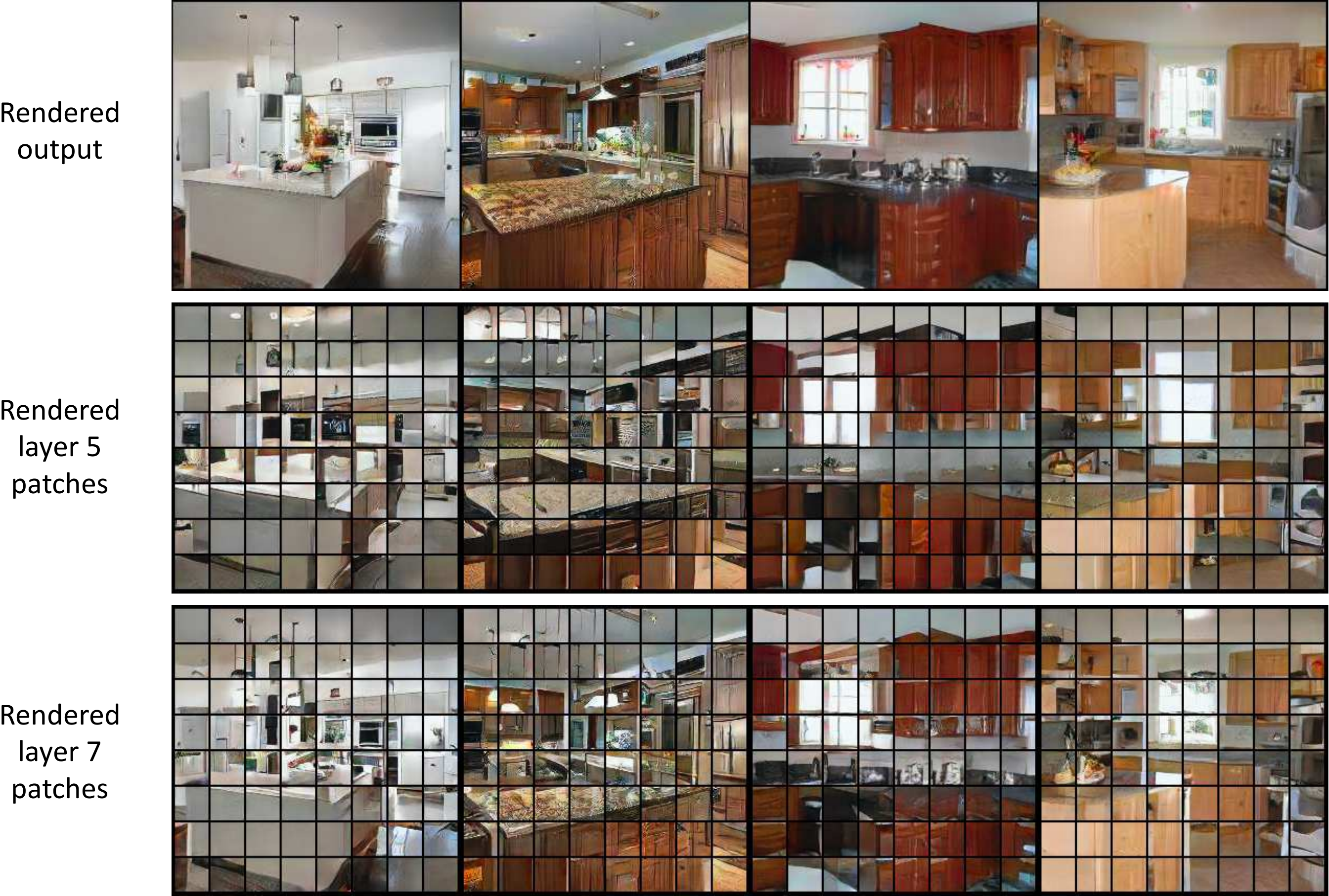}
    \caption{Comparison of rendered cropped activations at various layers of Progressive GAN generated LSUN kitchen images.}
    \label{fig:supp-proggan-kitchen-patch}
\end{figure}

Figures~\ref{fig:supp-stylegan-church-patch} and \ref{fig:supp-stylegan-kitchen-patch} visualize individual patches rendered in isolation at various layers of StyleGANv2, and compare those to the entire image rendered together.  Figures~\ref{fig:supp-proggan-church-patch} and   \ref{fig:supp-proggan-kitchen-patch}  visualize the same for Progressive GAN.


\end{document}